\documentclass[10pt,letterpaper]{article}
\usepackage{amsmath,amssymb,amsthm,booktabs,array,tabularx,graphicx}
\usepackage{mmla_technical_report}
\usepackage{orcidlink,placeins,needspace}
\graphicspath{{figures/}}
\newcommand{\E}{\mathbb E}
\newcommand{\R}{\mathbb R}
\newcommand{\sg}{\operatorname{sg}}
\newcommand{\TV}{\operatorname{TV}}
\newcommand{\cF}{\mathcal F}
\newcommand{\cG}{\mathcal G}
\newcommand{\cL}{\mathcal L}

\newcolumntype{Y}{>{\raggedright\arraybackslash}X}
\theoremstyle{definition}
\newtheorem{definition}{Definition}[section]
\theoremstyle{plain}
\newtheorem{proposition}[definition]{Proposition}
\newcounter{rsiAlgorithm}

\title{MMLA-RSI: Recursive Self-Improvement\\through RTT and RTMA}
\author{%
\textbf{Junyi Zou}\textsuperscript{1,*\,\orcidlink{0009-0009-1367-7428}}\quad\textbullet\quad
\textbf{Avrova Donz}\textsuperscript{1,2,*\,\orcidlink{0009-0009-0100-0719}}\\[-0.06em]
{\fontsize{7.3pt}{8.2pt}\selectfont \textsuperscript{1}MMLA-org\quad
\textsuperscript{2}Communication University of China (CUC), No. 1 Dingfuzhuang East Street, Chaoyang District, Beijing 100024, China}\\[-0.10em]
{\small \textsuperscript{*}Equal contribution}\\[-0.10em]
{\scriptsize Email:\enspace\href{mailto:lior.j.zou@gmail.com}{\nolinkurl{lior.j.zou@gmail.com}}
\enspace$\cdot$\enspace\href{mailto:avrovadonz@icloud.com}{\nolinkurl{avrovadonz@icloud.com}}}}

\date{October 14, 2026}
\hypersetup{pdftitle={MMLA-RSI: Recursive Self-Improvement through RTT and RTMA},pdfauthor={Junyi Zou and Avrova Donz},pdfsubject={Policy-state and authoritative-memory routes to recursive self-improvement},pdfkeywords={MMLA, recursive self-improvement, reasoning-time training, fast weights, causal evaluation}}
\begin{document}
\makemmlatitle{%
An agent can improve its behavior within an unresolved problem by changing how it acts and which experience it retains. We propose MMLA-RSI, a recursive self-improvement scheme with two implementations in Memory-Mediated Learning Architecture: Reasoning-Time Training (RTT) changes the policy state $\Phi$, while Reasoning-Time Memory Adaptation (RTMA) changes authoritative memory $M$. Each route also makes its adaptation method an object of improvement. A current numerical updater trains the policy proposing its successor; a current memory manager chooses the records available to the process proposing its successor. An independently admitted successor takes over the same role in the next generation.

\medskip\noindent We give bounded executable rule languages, shared task/proposal interfaces, and one algorithm for either route or their composition. The RTT instance changes scalar schedules in a clipped-momentum update. The RTMA instance chooses NULL or a complete-row write with possible eviction. A fixed interpreter, trusted committer, and independent evaluator bound what can change. State versions, read closures, and conditional properties describe causal execution, accepted transitions, and finite work.

\medskip\noindent Existing numerical-component studies motivate the construction. In four-candidate program selection, retaining LARC feedback updates reduces expected error by 24.65 and 36.65 percentage points for two trained initializations. In chronological continuous-integration replay, feedback binding is positive while retention increases loss. Completed SQL studies improve substantially over an unadapted substrate but do not establish the required additional online benefit. These results characterize their declared components. The two recursive routes and their joint realization are specified here; their beneficial recursive chains and discovery efficiency remain to be evaluated.}

\begin{center}
\begin{minipage}{.96\textwidth}
\includegraphics[width=\linewidth]{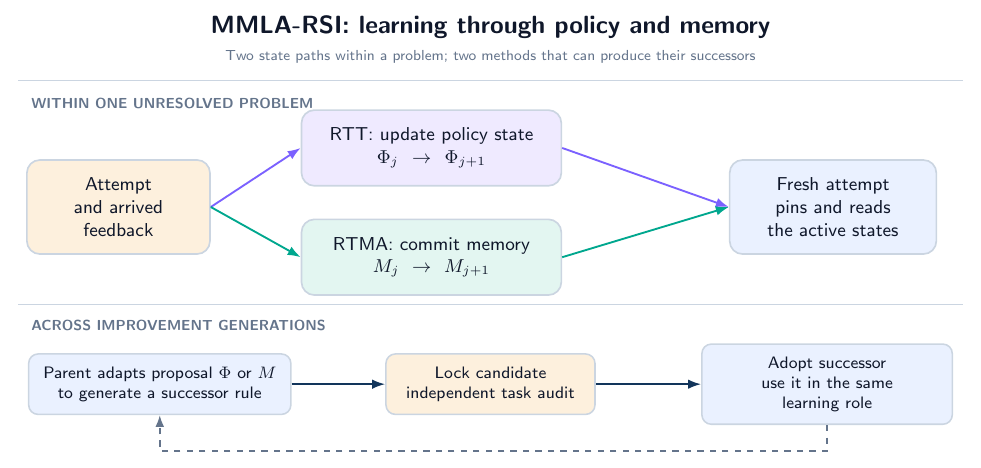}
\captionsetup{hypcap=false}
\captionof{figure}{\textbf{Two state paths, two recursive implementations.} RTT changes $\Phi$ through a numerical updater; RTMA changes $M$ through a memory manager. In each route, the current rule changes the proposal process that produces its successor. Admission installs that successor for the next generation. The routes can operate separately or share an unresolved problem.}

\label{fig:overview}
\end{minipage}
\end{center}
\clearpage

\section{Introduction}
An unsuccessful attempt supplies information that the next attempt may need. A program fails an execution check, a query violates a visible constraint, or a candidate learning rule performs poorly on a development task. The agent must decide both how this feedback changes its behavior and which parts remain available later. Memory-Mediated Learning Architecture (MMLA) makes these two learning objects explicit: policy state $\Phi$ and authoritative memory $M$~\cite{mmla}.

Reasoning-Time Training (RTT) updates $\Phi$ while a problem remains unresolved, then uses that state in a later attempt. Reasoning-Time Memory Adaptation (RTMA) commits a non-NULL successor of $M$ that a later attempt reads. The distinction follows the MMLA v4 state account. Here both paths are central. Learning a useful memory is one level of adaptation; improving the method that selects and maintains that memory is another. The same distinction holds between updating a policy state and improving its updater.

MMLA-RSI connects each adaptation method to the process producing its own successor. In the RTT implementation, the current numerical updater trains a private proposal policy. In the RTMA implementation, the current manager chooses the authoritative records that the proposal process can read. A locked successor is evaluated independently and, if admitted, replaces its parent in the same proposal-learning role:
\begin{align}
 U_{\eta_k^\Phi}&\longrightarrow\Phi^{\rm prop}_{j+1}
 \longrightarrow\eta_{k+1}^\Phi
 \longrightarrow\text{next proposal learning with }U_{\eta_{k+1}^\Phi},\nonumber\\
 V_{\eta_k^M}&\longrightarrow M^{\rm prop}_{j+1}
 \longrightarrow\eta_{k+1}^M
 \longrightarrow\text{next proposal learning with }V_{\eta_{k+1}^M}.
 \label{eq:route}
\end{align}
The middle arrows include generation, selection, and independent admission. The inherited objects are the rules $\eta^\Phi$ and $\eta^M$; proposal states may restart from fixed initializations each generation. A joint instance carries a pair of rules through the same cycle. Changing a state alone is adaptation. A recursive improvement chain additionally requires the changed adaptation method to participate functionally in producing a later improved method.

This dependency has direct precedents. Self-referential meta learning studies systems that modify their learning behavior~\cite{selfreferential}; STOP executes an improver on itself, and Darwin G\"odel Machine uses descendant agents in further self-modification~\cite{stop,dgm}. Our construction specifies the dependency through MMLA's two state paths: bounded numerical update programs and bounded memory-management programs, their task/proposal interfaces, and independent admission before role replacement. It makes both the changed object and its later use inspectable.

MMLA is mixer-agnostic. Low-Rank Adaptive Residual Connections (LARC) provide the current numerical carrier in front of a reused frozen mixer~\cite{larc}. The Transformer mixer and low-rank factors are implementation choices. $\Phi$ denotes a policy-state role, not a synonym for adapter parameters; another declared fast state can fill it. $M$ is the committed authority for retained records, not an undifferentiated name for prompts, logs, or caches. These distinctions let the same architecture support policy adaptation, memory adaptation, and their composition.

The method contribution is a complete recursive construction with two concrete bounded domains. The RTT domain varies feedback-dependent schedules in a clipped-momentum template. The RTMA domain varies the rule selecting NULL or a complete-row write under finite memory capacity. We specify their state, feedback, failure, read, and version semantics, then give a common algorithm and its conditional properties. Existing component studies motivate separate measurements of functional participation, task benefit, and discovery efficiency. The proposed method, an observed beneficial recursive chain, and an advantage over fixed search are distinct levels of evidence.

\section{The MMLA State, RTT, and RTMA}
\label{sec:rtt}
\subsection{Three timescales and two learning objects}
It is helpful to separate an attempt, a problem, and an improvement generation. Attempt $j$ produces a proposal for a fixed problem $(X,I)$, where $X$ is its input and $I$ is its immutable task identity. Feedback from that attempt may change the private state before attempt $j+1$. An improvement generation $k$ binds an adaptation-rule version $\eta_k$ across a collection of such problems. It can end by installing a new version for future problems.

The process state extends the state account in the MMLA report:
\begin{equation}
 S_j=(X,I,C_j,M_j,W_j,\Phi_j,\Xi_j^{\rm read},\Xi_j^{\rm aux},b_j),
 \qquad \Psi_j^\Phi=(\Phi_j,\Xi_j^{\rm read,\Phi}).
 \label{eq:state}
\end{equation}
Here $C$ is working context, $M$ is authoritative memory, $W$ is external world state, and $b$ is remaining resource budget. $\Xi^{\rm read}$ contains auxiliary state capable of affecting a later action, including an optimizer when it participates in a subsequent update, a random-number state, a tool cache, or a pending callback. $\Xi^{\rm aux}$ contains records that cannot influence the measured continuation. Calling a log ``auxiliary'' requires establishing that the continuation cannot read it.

The corresponding memory closure is $\Psi^M=(M,\Xi^{\rm read,M})$. The joint closure includes the union of dependencies, counted once when shared. Each controlled kernel below uses the appropriate declared closure $\Psi$. Slow initialization $\rho$ and initial memory $M_0$ say where a learner begins; $\eta^\Phi$ and $\eta^M$ say how feedback changes it. A run fixes its mutable domain to the RTT rule, the RTMA rule, or their pair. The pretrained substrate, language interpreters, memory authority checks, and evaluator remain bound outside that domain.

\begin{table}[htbp]
\centering\small
\begin{tabularx}{\linewidth}{@{}lYY@{}}
\toprule Object & Operation and lifetime & Restoration or transition \\
\midrule
$\theta_0,\omega$ & Bound pretrained substrate and shared offline adaptation & Verify original identity; never recreate by retraining. \\
$\rho$ & Slow initialization of the numerical carrier & Copied at problem start; bound for each comparison. \\
$\Phi$ & Private state changed by feedback in the current problem & Reset to the problem's bound $\rho$. \\
$\eta^\Phi,\eta^M$ & Numerical updater and memory-management rule & The selected domain changes only by admitted version replacement. \\
$M$ & Bounded authoritative records with verified complete-row commits & Starts from bound $M_0$; independently retained, reset, or held fixed. \\
$\Xi^{\rm read}$ & Every additional dependency of a future action & Restored with the intervention's declared closure. \\
$W,b$ & External effects and resource consumption & Effects and consumed resources survive a rollback. \\
\bottomrule
\end{tabularx}
\caption{Ownership and lifetime. Policy learning, memory learning, and changes to their adaptation methods are distinct operations.}
\label{tab:states}
\end{table}

\subsection{Feedback must arrive before it can teach}
Let an attempt finish at $\epsilon_j$, its feedback become visible at $v_j$, and an update commit at $u_j$. Let $\sigma_{j+1}$ be the start of the next fresh proposal and $\tau$ the resolution time of the current problem. The required event order is
\begin{equation}
 \epsilon_j\leq v_j\leq u_j<\sigma_{j+1}<\tau.
 \label{eq:timing}
\end{equation}
An update after resolution may improve another problem, but it does not establish the within-problem RTT or RTMA path for this one. Reranking candidates generated before an update does not show that their content was generated by the updated state. A newly computed selection action can nevertheless depend functionally on that state; its action space is candidate selection rather than candidate-content generation.

Let $(\cF_t)_t$ be the filtration of legally available information. The updater has the form
\begin{equation}
 (\Phi_{j+1},s_{j+1})=\Pi_{\mathcal K(b_j)}
 U_{\eta_k^\Phi}(\Phi_j,s_j,r_j;\zeta_j),
 \qquad r_j\text{ is }\cF_{v_j}\text{-measurable},
 \label{eq:update}
\end{equation}
where $r_j$ is the arrived record, $s_j$ is private update state, $\zeta_j$ is permitted randomness, and $\Pi_{\mathcal K(b_j)}$ enforces the declared feasible update set. The record can contain an attempt, its execution response, and a derived gradient. It cannot contain the answer to a later independent evaluation. An offline teacher can supply targets when training $\rho$; its presence must not be mistaken for online feedback available to $U_\eta$ at deployment.

\begin{figure}[htbp]
\centering\includegraphics[width=\linewidth]{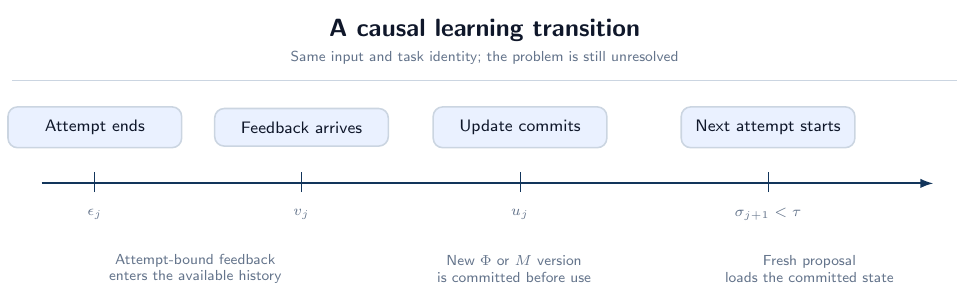}
\caption{\textbf{The shared timing condition.} The unresolved task persists from attempt to feedback, commit, and a later pinned read. RTT commits a policy-state update; RTMA commits a non-NULL memory successor. Independent future audit outcomes remain outside both paths.}
\label{fig:timeline}
\end{figure}

The subsequent action kernel averages over a declared source of continuation randomness. To distinguish this object from a replay, write $F_{\bar\Psi}(h,Z)$ for the executed action, where $h$ fixes the reached task history and arrived inputs, $\bar\Psi$ fixes the non-random read state, and $Z\sim\nu$ is the continuation random tape. RNG components of $\Xi^{\rm read}$ are conditioned on in a replay or marginalized under the declared law in a policy comparison; the same random tape is not simultaneously fixed and marginalized. Thus
\[
 K(A\mid h,\bar\Psi)=\int \mathbf1\{F_{\bar\Psi}(h,z)\in A\}\,\nu(dz).
\]
At a reached history, functional use in the marginal action-distribution sense requires
\begin{equation}
 \TV\!\left(K(\cdot\mid h,\bar\Psi_{j+1}),K(\cdot\mid h,\bar\Psi_j)\right)>0,
 \label{eq:functional}
\end{equation}
with non-target inputs and the random-tape law held fixed. The action space must also be declared: a token, a complete proposal, and a selected program define different kernels.

A common-random-number replay compares the conditional execution mappings $F_{\bar\Psi_{j+1}}(h,z)$ and $F_{\bar\Psi_j}(h,z)$. A changed output witnesses sensitivity at that fixed tape. It does not in general prove Equation~\ref{eq:functional}, because for this coupling
\[
 \TV(K_{j+1},K_j)\ \leq\
 \Pr_{Z\sim\nu}\!\left[F_{\bar\Psi_{j+1}}(h,Z)\ne F_{\bar\Psi_j}(h,Z)\right].
\]
Thus replay disagreement alone does not establish a marginal difference.\footnote{For a fair binary $Z$, the maps $F_0(Z)=Z$ and $F_1(Z)=1-Z$ disagree on every coupled replay while both action distributions are fair.} When the evaluated configuration is deterministic and $\nu$ is a point mass, different outputs do prove a difference between those conditional kernels.

Evidence for the marginal criterion can instead compare exact action probabilities on a finite action space, or a fixed measurable event whose probability differs with appropriate uncertainty. At a fixed generation prefix, a changed next-token probability is one such event for the corresponding continuation kernel. Replay sensitivity, estimated distributional dependence, and exact distributional dependence are reported under their respective objects. The saved SQL output changes below remain replay observations.

\subsection{What an RTT witness must record}
Table~\ref{tab:witness} summarizes the parent report's eight obligations for an inspectable RTT transition~\cite{mmla}.
\begin{table}[htbp]
\centering\small
\begin{tabularx}{\linewidth}{@{}lY@{}}
\toprule Witness & Required evidence \\
\midrule
W1 & The input and task identity remain the same and the problem is unresolved. \\
W2 & Feedback is bound to a completed attempt and has arrived before the update. \\
W3 & A genuine learning-state change is committed, with a version and read closure. \\
W4 & The update is causal, belongs to the current epoch, and does not use future information. \\
W5 & The commit precedes a fresh proposal and both occur before resolution. \\
W6 & The next proposal actually loads the committed state version. \\
W7 & A controlled intervention shows functional action-kernel dependence at a reached history. \\
W8 & Update, read, cost, lifetime, reset, and late-callback behavior are recorded. \\
\bottomrule
\end{tabularx}
\caption{The RTT witness. W1--W6 and W8 establish the process; W7 adds functional use. Task benefit is a separate empirical quantity.}
\label{tab:witness}
\end{table}
\FloatBarrier

The distinction between functional use and benefit is essential. An update can change the next output and make it worse. Real feedback can be less damaging than permuted feedback while both are worse than resetting. Conversely, a strong starting policy can solve a task without requiring a useful fast update. Section~\ref{sec:evidence} contains examples of all three issues in the component research program.

\subsection{Authoritative memory and the RTMA path}
The memory path follows MMLA v4\textquotesingle s definitions of RTMA and dual-state composition~\cite[Section~72, Definitions~72.2--72.3]{mmla}. Its transition separates a proposed action from an authorized commit:
\begin{equation}
 a_j^M=V_{\eta_k^M}(M_j,d_j),\qquad
 (M_{j+1},v^M_{j+1})=\operatorname{Commit}(M_j,a_j^M,d_j).
 \label{eq:memory-transition}
\end{equation}
The fixed assembler constructs the candidate record $d_j$ from arrived evidence. The manager chooses an allowed action. The trusted committer checks its type, provenance, capacity, expected predecessor, and epoch before installing a complete successor. NULL leaves authoritative content and its version unchanged and produces an explicit no-op receipt. In the minimal instance, a non-NULL action writes one complete row into a selected slot, evicting its previous row if occupied. The later attempt pins the actual successor version. A failed pin, an unread write, and a NULL receipt do not establish the positive RTMA path.

Authority concerns which committed records the reader treats as its stored evidence. It does not turn a stored hypothesis into a fact. A row can authoritatively record that an attempt received a particular execution response, with the response's scope and provenance, without claiming the attempted answer was correct. The manager cannot rewrite the evaluator, forge provenance, or enlarge its own access permissions.

The two paths have independent receipts. RTT requires the policy update/read and functional-use obligations in Table~\ref{tab:witness}. RTMA requires a non-NULL complete-row commit and later same-problem read; functional memory dependence is additionally tested by intervening on the memory closure with $\Phi$ fixed. A joint interval includes both paths and their interleaving. Neither classification implies positive task benefit.

A KV cache belongs to the state role determined by its production and use. A cache deterministically rebuilt from an unchanged transcript is working context or a read cache. An explicitly updated, versioned policy-state carrier may serve as $\Phi$ if it satisfies the declared update/read semantics. It is not authoritative $M$ merely because it stores tensors, and changing it does not automatically establish RTT.

\section{Numerical Carrier and Initialization}
\label{sec:carrier}
\subsection{The RTT carrier: a slow initialization and a private fast residual}
The present carrier inserts a low-rank residual before a frozen mixer:
\begin{equation}
 T_\Phi(h)=h+B_\Phi A_\Phi h,\quad
 A_\Phi\in\R^{r\times d},\quad B_\Phi\in\R^{d\times r},\quad
 f_{\theta,\Phi}(x)=F_\theta\bigl(T_\Phi(E_\theta(x))\bigr).
 \label{eq:larc}
\end{equation}
The factorization follows LoRA~\cite{lora}; the input-side placement, slow/fast state lifetime, and feedback experiments are described in the LARC report~\cite{larc}. With $d=1536$ and $r=4$, the two factors have 12,288 entries. Random nonzero $A$ and zero $B$ give an initial zero residual. Setting both factors to zero would remove the first task-gradient path to both factors. At the intended initialization, the first task gradient of $A$ is zero while the gradient of $B$ can be nonzero.

After slow training, $\rho=(A_\rho,B_\rho,v_\rho)$ is the starting state, where $v_\rho$ identifies the initialization version. A new problem copies these factors into $\Phi$. Reset restores this trained state, rather than a newly randomized $A$ or a bare zero $B$. A complete resume additionally restores the bound updater, private optimizer, random streams, and input cursor. Equal products $BA$ need not imply equal subsequent factor-space updates; storing the product alone therefore cannot reproduce a learning trajectory.

Frozen weights still participate in the derivative with respect to an input residual. Wrapping the entire downstream mixer in a no-gradient region would break this path. Similarly, a cached mixer output generally depends on $\Phi$ even though the mixer weights are frozen. Only representations before the first $\Phi$-dependent operation can be reused without including the fast-state version in the cache identity. Batched problems share the same slow version but own separate fast factors, optimizer state, and random streams; attention masks and positions must preserve those boundaries.

\subsection{Two ways to train the starting state}
For an episode with support $S$ and an outer-training query $Q$, let $\Phi^{(0)}=\rho$ and let $\Phi^{(m)}=U_\eta^m(\rho;S)$. Two initialization objectives are
\begin{align}
 J_{\rm static}(\rho)&=\E_{S,Q}\,\cL_Q(\rho),\\
 J_{\rm adapted}(\rho)&=\E_{S,Q}\,\cL_Q\bigl(U_\eta^m(\rho;S)\bigr).
 \label{eq:objectives}
\end{align}
In the paired program-selection experiment, both arms have the same residual placement, support, query supervision, and three paired initializations. The adapted arm uses a first-order MAML-style estimator~\cite{maml,fomaml}:
\begin{equation}
 \widehat g_\rho=\left.\nabla_\Phi\cL_Q(\Phi)\right|_{\Phi=U_\eta^m(\rho;S)}.
 \label{eq:fomaml}
\end{equation}
This assigns the adapted query gradient to corresponding slow factors, approximating the inner-update Jacobian by the identity. It is neither differentiation through the exact full inner trajectory nor a claim that a detached fast gradient automatically updates $\rho$. The implementation must explicitly aggregate episode gradients and apply them to the bound outer parameters.

Learning $\rho$ with a fixed $U_\eta$ prepares an adaptable starting point. It does not change the updater. Even learning $\eta$ using a fixed external meta-optimizer need not create the recursive dependency in Equation~\ref{eq:route}. The proposed next layer uses the current $U_\eta$ in the process that proposes its replacement. Keeping these objects separate prevents an ordinary initialization experiment from being counted as an experiment on recursive improvement.

\section{Measuring Feedback and Retention}
\label{sec:causal}
\subsection{The complete read closure}
A clean intervention specifies everything that can reach the measured continuation. Removing feedback from the visible prompt is insufficient if a selector reads a feedback log, an optimizer resumes with feedback-dependent momentum, or an asynchronous tool response arrives after a reset. The read closure includes all such paths. For a $\Phi$-only comparison, every other causal input must be identical across the paired reads. If auxiliary state is changed together with $\Phi$, the estimand is the effect of the bundle $\Psi^\Phi$ in Equation~\ref{eq:state}.

The program-selection study uses a particularly narrow read: the next query sees the bound model and residual but no support-feedback text, inner optimizer, or post-update cache. Memory is disabled. Later SQL studies use broader task-matched reads, including shared support and feedback context and selectors. Their $\Phi$ effects are conditional on these channels. Disabling $M$ alone does not make such a continuation the memory-disabled, feedback-text-free read of the earlier study.

\begin{figure}[htbp]
\centering\includegraphics[width=\linewidth]{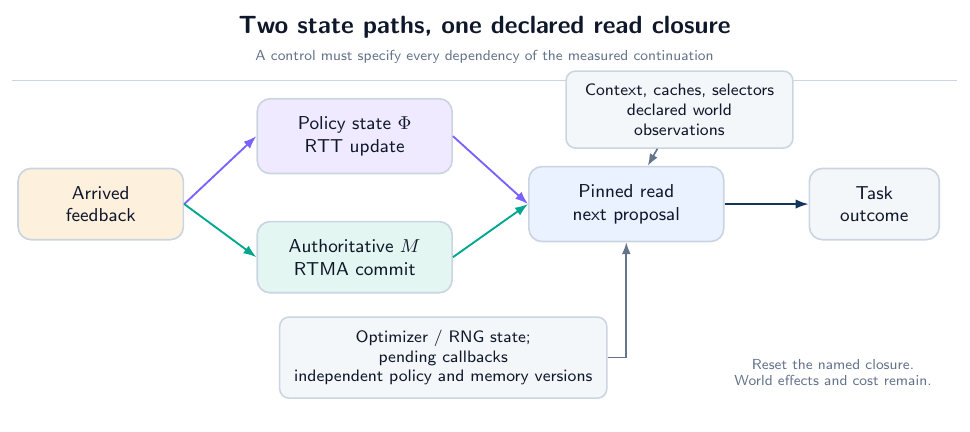}
\caption{\textbf{Two controlled state paths and their read closure.} Purple marks the policy path; teal marks authoritative memory. Context, caches, selectors, random streams, and callbacks are additional dependencies. Each intervention specifies its target and the other paths held fixed.}
\label{fig:closure}
\end{figure}

\subsection{Policy retention and feedback binding}
Let $L_{fk}$ be expected loss when feedback binding is $f\in\{\mathrm{real},\mathrm{sham}\}$ and retention is $k\in\{\mathrm{keep},\mathrm{reset}\}$. Sham feedback independently applies a uniformly sampled permutation over the declared eligible feedback positions. It retains the feedback values while breaking their binding; an identity permutation and an uninformative record remain legitimate draws. Both cells perform the registered updates. Reset occurs after those updates and before the paired continuation, restoring the arm's own $\rho$ and the specified closure.
\begin{align}
 G&=L_{\rm real,reset}-L_{\rm real,keep},\label{eq:G}\\
 D&=G-\left(L_{\rm sham,reset}-L_{\rm sham,keep}\right).\label{eq:D}
\end{align}
$G$ measures the value of retaining the real-feedback state relative to resetting. $D$ measures the additional retention effect attributable to the real binding. If both reset distributions coincide, $D=L_{\rm sham,keep}-L_{\rm real,keep}$. For a higher-is-better score, all loss contrasts are reversed consistently. Positive $D$ does not imply positive $G$.

Complete reset equality is a useful implementation consequence. If the restored causal state, future exogenous-input distribution, and action kernels are identical, the continuation distributions agree. Pathwise equality additionally requires the same coupled randomness and deterministic realization of the remaining operations. This equality concerns the declared continuation, not external world effects that an earlier action has already caused. A transaction can restore numerical state without undoing a sent request or refunding its computation.

\subsection{Memory effects and joint task effects}
For memory, compare a committed write with NULL from the same reached state, with the same arrived information and continuation budget. Recompute the complete future task outcome under each branch: the write can add useful information and remove useful old information. A memory reset restores the bound pre-intervention snapshot or $M_0$, as registered, together with its read caches and index versions. It is a different intervention from rejecting one proposed write. A memory-rule comparison gives an exact selector or strong fixed heuristic the same legal information, action set, capacity, and total computation allowance.

A joint evaluation uses policy keep/reset and memory keep/reset as separate factors, rather than relabeling the real/sham $\times$ policy-retention matrix. For the local retention probe defined here, branch from one common unresolved checkpoint after the registered feedback updates and before the next read. Keep uses that checkpoint's carrier state. Policy reset restores the bound $\rho$ and its initial private optimizer state; memory reset restores the task's bound $M_0$ and its index state. Invalidate or rebuild dependent caches for each resulting pair and verify its new joint pin. Hold the other reached inputs, external state, continuation random-tape law, and resource allowance fixed. Freeze further $\Phi$ and $M$ adaptation during the declared finite continuation and evaluate the same endpoint in all four cells. Let $L_{pm}$ be its loss with policy retention $p\in\{0,1\}$ and memory retention $m\in\{0,1\}$ under one declared feedback condition. Then
\[
 G_\Phi(m)=L_{0m}-L_{1m},\qquad G_M(p)=L_{p0}-L_{p1},\qquad
 I=L_{01}+L_{10}-L_{11}-L_{00}.
\]
These are state-retention effects conditional on the reached checkpoint. Resetting one carrier does not remove its earlier influence on the other carrier, context, or world. A probe that resumes both adaptation rules defines a different continuation estimand; its effects can propagate through subsequent cross-state updates. Disabling a route from initialization instead defines a complete-trajectory effect. Each comparison binds its own reset and continuation specification.

$I$ is the extra loss reduction of keeping both beyond the sum of their separate reductions; equivalently, $I=G_\Phi(1)-G_\Phi(0)=G_M(1)-G_M(0)$. Joint usefulness is judged by the complete task endpoint against same-information strong and frozen methods; neither two positive main effects nor a positive interaction is required merely to define the composed mechanism. Feedback-binding interventions can be added as a separately declared factor. The component values of $D$ below remain policy binding effects in their original read conditions, not estimates of $I$.

If $\Phi$, its updater, or the memory reader changes, a cached memory-value estimate may no longer describe the new continuation. Such caches must bind all causal reader identities and be invalidated or recomputed after a mismatch. Repricing, retrieval, write selection, eviction, and restoration enter the task cost. The minimal memory manager below uses directly observed row features rather than a learned value cache; a predictive value estimator is a possible larger domain, not a prerequisite for the RTMA construction.

\subsection{From a component effect to an RSI effect}
An adapted initialization can outperform another initialization without gaining anything from its own feedback. The objective increment and each arm's real-feedback increment therefore appear separately in the results. The recursive evaluation likewise reports parent participation and accepted task improvement separately from discovery efficiency; Section~\ref{sec:interventions} gives the corresponding interventions and comparisons.

Each comparison uses the original unit of generalization. Multiple seeds applied to one task group are repeated measurements within that group; they are not additional independent tasks. Selection uses development groups, and a locked candidate is evaluated on separate audit groups. Once an audit outcome influences a successor's design, that group becomes part of the development history for subsequent statistical claims. New accepted generations require fresh audit information under the stated admission assumptions.

\section{Two Recursive Implementations and Their Composition}
\label{sec:recursive}
Fix the mutable domain before a run: $\mathcal H_\Phi$ for RTT, $\mathcal H_M$ for RTMA, or $\mathcal H_\Phi\times\mathcal H_M$ for a joint instance. We use $\eta$ for the selected rule or pair. In every case the parent changes a state read by the process proposing its successor, and the admitted successor takes over that same role. The numerical and memory languages below instantiate this construction with finite, explicit operations.

\subsection{RTT: a bounded numerical updater}
In the next three RTT subsections, $\eta$ abbreviates $\eta^\Phi$ and $\mathcal H$ abbreviates $\mathcal H_\Phi$. The updater is a bounded program with a typed input and output:
\begin{equation}
 (\Delta\Phi_j,s_{j+1})=\operatorname{Eval}_{\mathcal H}
 (\eta;\Phi_j,s_j,g_j,z_j),\qquad
 \Phi_{j+1}=\Pi_{\mathcal K}(\Phi_j+\Delta\Phi_j).
 \label{eq:dsl}
\end{equation}
$g_j$ is a feedback-derived gradient and $z_j$ contains arrived scalar statistics. The minimal instance below varies feedback-dependent step sizes, gradient-clipping thresholds, and momentum coefficients within a fixed clipped-momentum template. The state layout, gradient input, projection, and interpreter remain fixed. This gives a concrete bounded domain for the recursive construction.

Flatten the chosen LARC factors in row-major order, first $A$ and then $B$, into $\Phi\in\R^q$. The language's private state is one momentum vector $s\in\R^q$ in the same coordinate order, initially zero. A program is the canonical prefix serialization of three scalar expression trees, $\eta=(e_a,e_\kappa,e_\mu)$. Each tree has the grammar
\[
 e ::= v\mid z_i\mid (e+e)\mid(e-e)\mid(e\,e)
       \mid\min(e,e)\mid\max(e,e)\mid\operatorname{sdiv}_\varepsilon(e,e),
\]
where $v$ belongs to a fixed finite set $\mathcal V$ of finite FP32 constants containing $0$ and $1$, and $i\in\{1,\ldots,4\}$. Constants are serialized as indices into a bound ordered table of their exact FP32 bit patterns; operator names, delimiters, and index encodings are fixed. There are no free-form numerical literals. The product $e\,e$ denotes scalar multiplication. Safe division is
\[
 \operatorname{sdiv}_\varepsilon(x,y)=
 x/\bigl(\operatorname{sign}_+(y)\max\{|y|,\varepsilon\}\bigr),
 \quad \operatorname{sign}_+(0)=1.
\]
The interpreter traverses trees in fixed left-to-right postorder. All scalar and vector operations, including norm reduction order and rounding, use the bound numerical implementation. Every intermediate is checked for finiteness. The constructor bounds total syntax nodes by $N_{\rm AST}\geq4$, each expression depth by $D\geq0$, and the tokenized serialization length by $L_{\rm AST}\geq1$. Admission requires a nonempty grammar satisfying all three limits and containing the parent.

Let $\ell_j\in[0,1]$ be the arrived feedback loss. Each learner starts with $j=0$ and $\ell_{-1}=0$. Here $j$ indexes feedback-bearing attempts, including invalid or incomplete actions assigned a failure loss; it does not count only successful numerical updates. The task runner binds a cap $T^{\rm task}\geq1$ per task, while proposal learning uses $T^{\rm prop}=N$ attempts per generation and at most $N-1$ updates. The same fixed statistics map uses the current role's $T$:
\[
 z_j=\left(
 \ell_j,\ \operatorname{clip}_{[-1,1]}(\ell_j-\ell_{j-1}),\
 \min\{1,\|g_j\|_2/G\},\ \min\{1,j/T\}
 \right),\qquad G>0.
\]
The loss history and opportunity index belong to each learner's arrived record and reset with that learner. The updater cannot read raw task identities, future targets, memory, or evaluation internals. The frozen parser and bounds turn the three expressions into coefficients:
\[
 a_j=\operatorname{clip}_{[0,a_{\max}]}e_a(z_j),\quad
 \kappa_j=\operatorname{clip}_{[\kappa_{\min},\kappa_{\max}]}e_\kappa(z_j),\quad
 \mu_j=\operatorname{clip}_{[0,\mu_{\max}]}e_\mu(z_j).
\]
Here $0<a_{\max}<\infty$, $0<\kappa_{\min}\leq\kappa_{\max}<\infty$,
$0\leq\mu_{\max}<1$, and $\varepsilon>0$ are fixed algorithm parameters.
For $S_{\max},R_\Phi\in(0,\infty)$ the deterministic execution rule is
\begin{align*}
 \tilde g_j&=g_j\min\{1,\kappa_j/(\|g_j\|_2+\varepsilon)\},\\
 s^+&=\operatorname{clip}_{[-S_{\max},S_{\max}]}
       \bigl(\mu_j s_j+(1-\mu_j)\tilde g_j\bigr),\\
 \Delta\Phi_j&=-a_j s^+,\qquad
 \Phi^+=\operatorname{clip}_{[\rho-R_\Phi,\rho+R_\Phi]}
                  (\Phi_j+\Delta\Phi_j).
\end{align*}
Clipping of a vector is coordinatewise. $\rho$ is the bound starting vector for the current learner role. Evaluation commits $(\Phi^+,s^+)$ together. A nonfinite intermediate, malformed tree, exceeded instruction cap, or failed output check rejects the whole update and restores both $\Phi_j$ and $s_j$; neither tentative tensor can reach a later read. The deterministic updater consumes no random draws. Sampling and tool work already performed retain their RNG progression, observations, and charges. Reset of an episode and transactional rejection of one update are therefore different operations.

\subsection{RTT: one interface for task and proposal learning}
The task learner and proposal learner use the same factor layout and dimension $q$, with separately bound initializations $\rho^{\rm task}$ and $\rho^{\rm prop}$ and private momentum vectors. The same syntax tree is interpreted in both roles without a role-dependent branch. Only the gradient and arrived feedback come from different problems. No momentum or fast factors are shared between task executions or between the two roles.

For a complete minimal instantiation, fix a task runner that samples an action $y_j^{\rm task}$, receives a legal arrived feedback loss $\ell_j\in[0,1]$, and returns
\[
 g_j^{\rm task}=\sg(\ell_j)
        \nabla_{\Phi^{\rm task}}\log p_{\Phi^{\rm task}}(y_j^{\rm task}\mid h_j).
\]
This is the zero-baseline score-function estimate for minimizing that feedback loss; the probability is that of the actual sampling law. A task may instead bind a differentiable feedback loss, but that replacement defines a different fixed runner. The runner specifies its observation space, resolution condition, maximum $T^{\rm task}$ feedback opportunities, and final task loss in $[0,1]$. Invalid or incomplete actions receive the runner's fixed failure loss one and retain their actual sampling probabilities. These outcomes, ordinary validation failures, and tool-reported errors provide legal feedback: the runner continues its feedback/update cycle unless its resolution condition or $T^{\rm task}$ cap has been reached. Candidate updaters never receive the independent final audit answer as within-task feedback.

The proposal learner receives the development evaluation of a candidate program and uses Equation~\ref{eq:proposalgrad}. Its $\ell_j$ is the negative of the bounded proposal reward defined below. Both roles thus return the same typed tuple $(\Phi,s,g,z)$ to $U_\eta$. Type compatibility follows by construction; useful transfer of an updater between the two gradient distributions is an empirical question. The constant-tree program gives clipped momentum SGD, while feedback-dependent trees change its step size, clipping threshold, or momentum according to the arrived record. Its task utility is evaluated by the bound runner.

\subsection{RTT: how the updater participates in successor generation}
At generation $k$, form a proposal task: produce a legal updater program with better task performance under the common budget. Its input includes $\eta_k$, the fixed interface, and permitted development summaries. Begin a private proposal-policy state $\Phi^{\rm prop}_0=\rho^{\rm prop}$. The model samples a bounded token rollout $y_j\in\Omega_{\mathcal H}$ using a fixed grammar mask. This space includes terminated legal serializations and incomplete or invalid outcomes of the bounded sampler. Decoding returns a candidate $c_j\in\mathcal H$ or the failure symbol $\bot$. Candidate execution or its fixed failure branch produces an arrived bounded reward $r_j^{\rm prop}$. Algorithm~\ref{alg:recursive} specifies its value.

Candidate execution is discrete. We therefore use a score-function estimate~\cite{williams} for the proposal-policy loss:
\begin{equation}
 g_j^{\rm prop}=-\sg\!\left(r_j^{\rm prop}-\beta_j\right)
 \nabla_{\Phi^{\rm prop}}\log p_{\Phi^{\rm prop}}^{\mathcal H}
 (y_j\mid x_j).
 \label{eq:proposalgrad}
\end{equation}
The log probability is the probability of the actual sampled rollout $y_j$ under the declared grammar mask and stopping rule. It is not conditioned on successful parsing or renormalized over surviving candidates. A truncated rollout is scored through its actual sampled prefix; forced cap termination is deterministic. The failure reward therefore participates in the same bounded sampling objective. Equation~\ref{eq:proposalgrad} targets candidate development evaluation conditional on the pre-sample history; it is not an unbiased meta-gradient of the entire selection--audit--recursion trajectory. The reward is stopped from gradient propagation. Baseline $\beta_j$ is fixed from the pre-sample history or an independent sample; it must not covertly use the sampled candidate's own reward while claiming the usual unbiased score-function identity. Finite-precision estimates, masks, truncation, and reward clipping are part of the implementation identity.

Crucially, the proposal task then invokes its current updater:
\begin{equation}
 (\Phi^{\rm prop}_{j+1},s^{\rm prop}_{j+1})=
 U_{\eta_k}(\Phi^{\rm prop}_j,s^{\rm prop}_j,g_j^{\rm prop},z_j),
 \qquad y_{j+1}\sim p_{\Phi^{\rm prop}_{j+1}}^{\mathcal H}.
 \label{eq:proposalupdate}
\end{equation}
The fresh rollout $y_{j+1}$ and its decoded candidate are generated after feedback and the committed update, while the proposal task is unresolved. The transition records the actual factor and private-state changes, then the subsequent read. A zero current gradient can still produce an update when momentum is nonzero. Conversely, changed momentum alone need not change the next proposal if the reader uses only $\Phi$. Functional use is classified from the declared read and action kernel, not from the gradient norm or a version counter. A rejected update restores both factors and private updater state from a snapshot taken immediately before the updater call; already consumed sampling randomness, feedback records, and costs are retained.

Write the resulting generator as $\cG_\eta=\text{fixed proposal framework}\circ\text{RTT}(U_\eta)$. This expression says exactly what changes: the framework's code stays fixed, while the learning process it calls changes with $\eta$. The common typed interface makes the same update program executable in both learners. Whether it learns well on both gradient distributions is measured separately from its type and state compatibility.

\subsection{RTMA: a bounded memory-management program}
\label{sec:rtma-rule}
The memory route makes the management rule, rather than just the retained contents, a candidate for improvement. Its minimal mutable domain chooses which complete arrived evidence records to retain. It leaves record meaning, authority checks, representation, and reading fixed. No numerical gradient is required for this branch.

\paragraph{Records, capacity, and authority.}
Bind $S\geq1$ slots, a maximum serialized row size $B_M\geq1$, and an initially valid $M_0$ for each learner role. The authoritative capacity is at most $S B_M$ bytes, including row metadata and any stored derived view. A row records a bounded attempted action $y$, its arrived feedback loss $\ell\in[0,1]$, normalized charged cost $c\in[0,1]$, a completion/failure tag, an immutable source identifier, and the role-local opportunity index $t$. In the proposal role, $y$ is the actual sampled program rollout, including a failed or capped prefix; its evaluation loss is $-r^{\rm prop}$ and its cost is the development execution's $\bar C$. If parsing or bounded sampling produces no legal candidate, no development execution starts: set $\bar C=0$, loss one, and the explicit invalid-or-incomplete tag. Sampling, parsing, and failure handling still enter the total cost ledger. In the task role, $y$ is the sampled task action, $\ell$ is the runner's legal feedback loss, and $c$ is incurred attempt cost divided by its bound attempt allowance. All strings, integer encodings, tag vocabularies, token counts, and numeric representations are fixed and finite. There is no missing-value inference: malformed or oversized evidence is an operation failure; a legal failed attempt has an explicit tag and loss one.

The candidate payload $d$ is assembled once from those observations. For each target slot $i$, a fixed assembler constructs a complete replacement row $R_i(d;M)$, including slot and object identifiers, expected predecessor version, new row version, provenance binding, canonical payload, and its fixed-size derived view. The derived view may be a fixed encoding of the canonical fields; the bound deterministic encoder and consistency check are part of the carrier identity. The validator checks bytewise agreement with this encoding before publication and at verified reads. Reader access to the canonical view does not authorize omitting required integrity fields. Object identifiers are unique within the carrier, and an attempted duplicate insertion with incompatible lineage is rejected. The trusted validator determines eligible slots from the fixed protection and eviction rules; the minimal configuration marks all slots evictable. The manager cannot change those rules. Stored and temporary representations, the isolated archive, and construction costs are accounted for separately from resident capacity.

\paragraph{A shared scoring language.}
First validate the payload\textquotesingle s common schema, size, and provenance requirements. A malformed payload returns an operation failure with $M$ unchanged, before action features or scores are computed. A correctly formed record of a failed attempt remains a valid payload. Only after this common check passes are complete rows assembled and validated against the slot-specific constraints. A valid payload with no eligible write slot leaves NULL as the normal choice. The resulting legal action set is $\mathcal A(M,d)=\{\mathrm{NULL}\}\cup\{\mathrm{WRITE}_i: R_i(d;M)\text{ passes validation}\}$. An empty slot gives insertion; an occupied slot gives eviction followed by insertion. Let $o_i$ indicate occupancy and let $(\ell_i,c_i,t_i)$ be the current row's fields. Empty-slot fields are zero. NULL has $o_i=\ell_i=c_i=t_i=0$ and its age feature is zero. At opportunity $j$ with role cap $T$, define the fixed action features
\[
 f(M,d,a)=\left(
 \mathbf1_{a=\mathrm{NULL}},\ o_i,\ \ell,\ c,\ \ell_i,\ c_i,
 \operatorname{age}_i,\ \operatorname{occ}(M)/S
 \right),
 \quad
 \operatorname{age}_i=\begin{cases}
 \min\{1,(j-t_i)/T\},&a=\mathrm{WRITE}_i,\ o_i=1,\\
 0,&\text{otherwise}.
 \end{cases}
\]
Initial occupied rows have $t_i=0$; subsequent writes store the current $j$. Each learner starts at $j=0$, advances its read-only opportunity counter after a feedback cycle, and resets it at problem start. Time passing does not mutate rows. Features are in $[0,1]$ and have the same units in both roles, with $T=T^{\rm task}$ or $N$ and independently bound cost denominators. They include information about the proposed row and the displaced row, allowing the rule to trade new evidence against retained evidence. They are observations, not oracle estimates of future utility.

A manager $\eta^M=e_M$ is one scalar expression tree using the RTT grammar's finite constants and arithmetic operators, with leaves $f_1,\ldots,f_8$ instead of $z_1,\ldots,z_4$. Its own node, depth, instruction, and serialization bounds are fixed. The same finite-precision evaluation, safe division, and intermediate finiteness checks apply. The chosen action is
\begin{equation}
 V_{\eta^M}(M,d)=\underset{a\in\mathcal A(M,d)}{\operatorname{argmax}}
                  \ e_M\bigl(f(M,d,a)\bigr).
 \label{eq:memory-score}
\end{equation}
Ties use the fixed order NULL, then ascending slot index. Infeasible slots are omitted; NULL is always available. A scoring failure rejects the entire operation without changing $M$. A non-NULL selection is revalidated against the pinned predecessor and committed as a complete row; any failed validation is recorded as failure, not a silent NULL. The interpreter permits no I/O or access to payload text beyond the declared scalar features. It has no hidden recurrent state. The all-zero tree selects NULL; other trees can prefer an empty slot or replace an occupied high-loss record. These examples show the action-dependent domain, not an assumption that low-loss records are always most valuable.

NULL leaves all authoritative bits, versions, and indexes unchanged. Its receipt and cost go to the isolated execution record. A non-NULL commit changes the selected row and its authorized version/index metadata; it does not rewrite unrelated records. The later read is a verified pin of this successor. If all decisions are NULL, that run contains no positive RTMA memory path.

\paragraph{The memory seen by the next attempt.}
The fixed reader serializes every occupied canonical row in slot order, with unambiguous field delimiters and explicit evidence tags. It reads no omitted payload through an external pointer. A preflight check ensures the immutable header plus the maximum permitted memory serialization fits the model's context and read allowance. The header contains the current problem, parent program, and fixed interface, but no changing evaluation history. Each new attempt rebuilds its input and input-dependent KV cache from that header and its pinned $M$; evicted records cannot survive in an old cache. Archived attempts remain available to the final selector and auditor only. The proposal model and memory manager cannot read that archive. Cache keys include the pinned memory, policy, reader, and header identities when reuse is implemented.

\subsection{RTMA: learning memory and improving its manager}
\label{sec:rtma-recursion}
Task and proposal learners use the same row schema, action features, expression interpreter, and commit/read interface, with independent $M_0^{\rm task}$ and $M_0^{\rm prop}$. The task runner fixes the unresolved problem, legal action distribution, feedback, terminal endpoint, and opportunity limit. Between unresolved attempts it supplies a complete arrived payload to the active manager and then pins the resulting memory for the next attempt. In RTMA-only runs the policy state is fixed. The two roles share field semantics; whether one manager works well across their feedback distributions is evaluated, not inferred from type compatibility.

In generation $k$, the proposal problem is to produce a better manager within the fixed interface and budget. The model samples a candidate $c_j\in\mathcal H_M$ or an invalid bounded rollout. A legal candidate is evaluated as a manager in fresh task learners; the fixed development block returns its arrived loss and cost. The proposal learner then uses its \emph{current parent} to manage this evaluation record:
\begin{align}
 d_j^{\rm prop}&=\operatorname{AssemblePayload}(y_j,-r_j^{\rm prop},\bar C_j,\text{provenance}),\nonumber\\
 M_{j+1}^{\rm prop}&=\operatorname{Commit}\bigl(M_j^{\rm prop},
       V_{\eta_k^M}(M_j^{\rm prop},d_j^{\rm prop}),d_j^{\rm prop}\bigr),\nonumber\\
 y_{j+1}&\sim p_{\rho^{\rm prop}}^{\mathcal H_M}
       (\cdot\mid\text{header},\operatorname{Read}(M_{j+1}^{\rm prop})).
 \label{eq:memory-recursion}
\end{align}
The notation suppresses the commit receipt and version returned by Equation~\ref{eq:memory-transition}. The candidate's task executions end before their development summary arrives, but the enclosing proposal problem is still unresolved. Thus the write and next proposal occur in the same problem. Candidate managers never manage the current proposal memory before admission.

This route contains two learning levels. Within a generation, feedback changes the retained memory under $\eta_k^M$. Across generations, a candidate with independently improved task performance can replace that management rule. Generation $k+1$ resets proposal memory to the same bound $M_0^{\rm prop}$ and uses $\eta_{k+1}^M$ to choose the evidence for further proposals. The rule, rather than an ever-growing transcript, carries the cross-generation change. Merely storing a candidate's program in memory does not establish this dependency; the admitted program must execute as the next manager.

\subsection{A joint RTT--RTMA instance}
\label{sec:joint}
For the joint domain, a candidate is a canonical pair $(\eta^\Phi,\eta^M)$. Its task evaluation executes both rules. The proposal learner also uses both \emph{parent} rules on the same arrived development feedback: first compute the score-function gradient and commit the numerical update, then assemble the feedback payload and commit the selected memory action. Memory scores use only Equation~\ref{eq:memory-score}'s observed features; they do not secretly query the updated policy or a value cache. The next attempt pins the compatible pair $(v^\Phi,v^M)$ and reads both states. The sampler grammar, serialization cap, and candidate identity cover the whole rule pair.

Each carrier operation is atomic, but the pair is not a single transaction. A rejected numerical update restores its own factors and private momentum and terminates this minimal attempt before memory management. If it succeeds and memory selection returns NULL, retain the new $\Phi$ and unchanged $M$. If memory construction or commit fails, retain the already committed $\Phi$, preserve the receipts for the successful policy commit and failed memory operation, and end the task or generation without a further proposal. An incompatible version pair likewise prevents the next read; it does not undo successful commits. Costs, feedback already observed, and consumed random streams remain recorded in every case.

Admission of a joint candidate targets the task risk of the rule pair $(\eta^\Phi,\eta^M)$. It certifies neither component-wise improvement nor a beneficial interaction, and an accepted edge need not change both rule components. The joint construction does not require both states to change after every feedback event. Its positive dual-path evidence requires actual policy and non-NULL memory paths with later reads. Each route can affect complete task utility conditionally on the other; their individual effects and interaction are measured using Section~\ref{sec:causal}, rather than inferred from the fact that two modules ran.

\subsection{Lock, audit, adopt, and use}
After a finite proposal budget, choose at most the registered number of candidates using development evidence and lock their exact identities. Each locked candidate and its parent run on paired, previously unused audit tasks with the same starting state, information, and resource envelope. A fixed evaluator measures the primary task endpoint and cost. Its answers cannot update the proposal state that produced the candidate under audit.

If a candidate satisfies the fixed admission rule, commit $\eta_{k+1}$ and begin the next generation with a fresh proposal state. That next generation uses the admitted $U_{\eta^\Phi_{k+1}}$, $V_{\eta^M_{k+1}}$, or their pair in the same proposal-learning role. Otherwise retain $\eta_k$ or terminate according to the registered finite contract. A failed candidate remains in the record, with its development, audit, serialization, and restoration costs. Restoring a previous version is a state transition; it is not a new budget allowance.

\begin{figure}[htbp]
\centering\includegraphics[width=\linewidth]{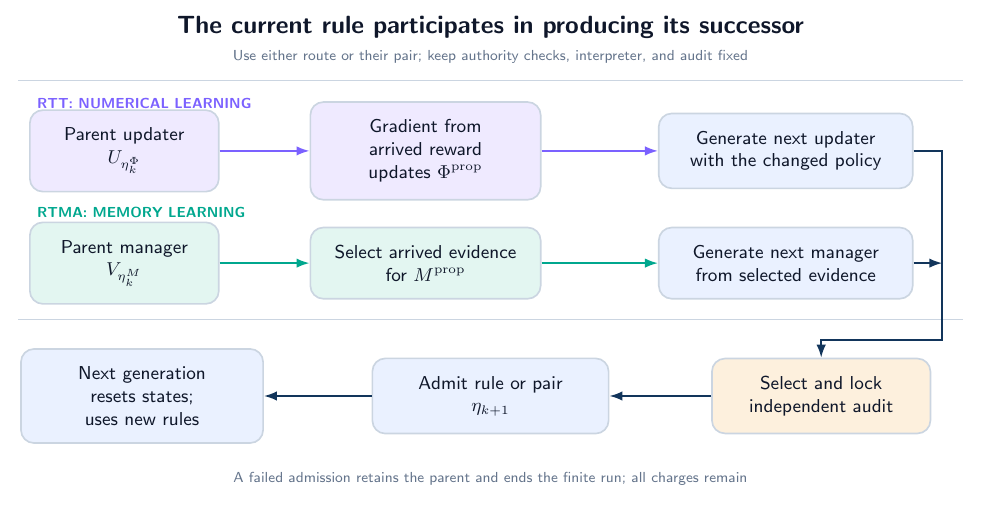}
\caption{\textbf{Two realizations of parent participation.} The RTT parent updates the numerical proposal state. The RTMA parent manages the authoritative evidence read by the next proposal. Selection and independent audit are shared; an admitted rule returns to its own role in the next generation. Joint runs carry both rules.}
\label{fig:recursion}
\end{figure}

\subsection{The complete recursive algorithm}
\label{sec:algorithm}
Algorithm~\ref{alg:recursive} collects the construction into one execution order. Its parameters fix a finite run without fixing experimental hyperparameters in this report. In addition to the language bounds, they include a generation cap $K\geq1$, a proposal cap $N\geq2$ per generation, a nonempty finite development task block, fresh audit blocks, the task runner, a cost weight $\lambda\in[0,\infty)$, a deterministic tie order, and finite resource reservations for every call. At most one candidate is locked and audited in each generation. Failure of admission ends this minimal run.

For a candidate $c$, execute the fixed task runner with its rule or rule pair from private copies of $\rho^{\rm task}$ and $M_0^{\rm task}$ on the development block. Return its mean final loss $\bar L(c)\in[0,1]$ and normalized charged cost $\bar C(c)\in[0,1]$, using the fixed per-execution cost allowance as denominator. For a legal program, a runtime exception that prevents legal feedback, a rejected numerical update or malformed memory operation, or exhaustion of a task-local allowance is a terminal task failure. It ends only that task with final loss one and its actual incurred cost; ordinary action, validation, and tool-error feedback follows the continuing runner described above. Evaluation then continues with fresh state on the remaining fixed tasks; means use the entire fixed block, never only completed successes. The same rule applies within an audit block. Exhausting the global reservation stops the generation without admission. The proposal reward is
\[
 r^{\rm prop}(c)=-\min\{1,\bar L(c)+\lambda\bar C(c)\}.
\]
This reward intentionally caps the combined loss-and-cost penalty at the unit failure level. Candidates above that level have the same proposal reward, so their within-region quality differences do not affect the score-function reward; the recorded cost still breaks selection ties. This is the chosen saturated objective, not an order-preserving normalization of the uncapped sum. The saturated reward guides development proposal learning and selection. Admission instead tests the bound task risk $R(\eta)$ in Equation~\ref{eq:admit} together with deterministic resource checks; a better development reward alone does not admit a successor. The minimal instantiation fixes $\beta_j=0$ in Equation~\ref{eq:proposalgrad}. The development task block and task random tapes are paired across candidates; each execution begins from fresh private state. Audit tasks are disjoint from development and from prior audits. A bound stream table assigns independent tapes to proposal sampling and task execution; paired candidate--parent audit runs share task tapes, not learner state.

The proposal input $x_j$ is the immutable header followed by the fixed read of $M_j^{\rm prop}$. For RTT-only runs memory stays at $M_0^{\rm prop}$; feedback affects later proposals through $\Phi$. For RTMA-only runs $\Phi$ stays at $\rho^{\rm prop}$; feedback reaches them through the selected memory rows. Joint runs expose both paths. This common input contract has no additional full candidate-history prompt. The bounded archive used for final selection is inaccessible to the generator and its adaptation rules. Fixed header and row bounds must fit the model context without truncating a committed row.

Generation uses the fixed grammar mask, temperature, token cap, and normalized sampling law for the selected domain. Initialize each attempt record with loss one, development cost $\bar C_j=0$, its source identifier, and an invalid-or-incomplete tag. A legal proposal executes the full fixed block and replaces these defaults with its measured loss, normalized cost, and completed-evaluation tag, including task-failure indicators. Each terminated legal proposal is scored once; incomplete or illegal output retains reward $-1$ and is ineligible for admission. Its actual bounded rollout can still supply RTT's score-function loss or RTMA's record of a failed attempt. Parsing does not renormalize the sampling law over successful candidates.

\begin{center}
\begin{minipage}{.97\linewidth}
\hrule\vspace{6pt}
\refstepcounter{rsiAlgorithm}
\textbf{Algorithm \thersiAlgorithm: MMLA-RSI with one locked successor per generation}
\label{alg:recursive}
\smallskip

\textbf{Inputs.} A domain $b\in\{\mathrm{RTT},\mathrm{RTMA},\mathrm{joint}\}$ and legal $\eta_0$; bound substrate, role initializations and memories; languages, typed record schema, reader and committer; task/development/audit partitions; stream table; $K,N$; fixed cost and admission rules.

\begin{enumerate}[leftmargin=1.6em,itemsep=4pt]
\item \textbf{Bind and reserve.} Check identities, partitions, language membership, context/capacity bounds, numerical implementation, and the finite ledger. Set $k=0$ and the admitted-version list to $[\eta_0]$.
\item \textbf{Start generation $k<K$.} Reserve its envelope, including one complete paired audit and closure work. Bind the parent rule or pair. Set $j=0$, $\Phi^{\rm prop}_0=\rho^{\rm prop}$, $s^{\rm prop}_0=0$, $M^{\rm prop}=M_0^{\rm prop}$, and $\ell_{-1}^{\rm prop}=0$. Bind $T^{\rm prop}=N$, initialize the isolated selection archive, and keep the proposal problem unresolved.
\item \textbf{Generate attempt $j<N$.} Pin compatible state versions. Sample a bounded rollout from the declared header and memory read with the current $\Phi$. Record its probability, parent and read versions, random-tape reference, and times.
\item \textbf{Obtain development feedback.} Initialize the typed attempt record as above, then decode and check the candidate. A legal candidate runs in fresh private task learners and replaces the defaults with its block results. Aggregate task-local failures as specified; invalid or incomplete proposals retain reward $-1$ and development cost zero. Charge all actual work to the ledger. Append the record to the isolated selection archive.
\item \textbf{Adapt before another attempt.} If $j+1=N$, go to step 6. Otherwise invoke the \emph{parent}, not the candidate: RTT computes Equation~\ref{eq:proposalgrad} and calls $U_{\eta_k^\Phi}$; RTMA assembles $d_j$ and calls $V_{\eta_k^M}$ plus the trusted committer; joint executes RTT then RTMA from the same arrived feedback. Record actual changes, NULL/rejection receipts, and the next read versions. On a failed operation, end the generation without admission using the rollback rules in the text. Otherwise advance the preceding-loss record and $j$, then return to step 3.
\item \textbf{Lock one candidate.} Eligibility requires a preceding committed parent numerical update for RTT, a preceding non-NULL parent memory commit for RTMA, and both for joint; subsequent attempts must read their recorded descendant states. Among eligible legal proposals choose the highest development reward, lowest cost, then lexicographically smallest canonical program; duplicates use the earliest eligible attempt. If none exists, terminate. Lock the candidate, settings, and selection provenance.
\item \textbf{Audit and decide.} Run the candidate and parent on paired fresh audit tasks from their bound task states. Apply Equation~\ref{eq:admit} and resource checks. Audit outcomes do not enter this generation's proposal state. An incomplete audit or failed admission terminates with the parent retained.
\item \textbf{Adopt and recur.} Commit the admitted rule or pair $\eta_{k+1}$ and its version edge. Close the problem, release private state, and increment $k$. If $k<K$, return to step 2: reset the proposal states but use the new rule in the same adaptation role.
\item \textbf{Deliver on every exit.} Save the last admitted version, all candidates and failures, both state/read receipt chains, selection/audit identities, and cumulative charges. Exhaustion or absence of a successor ends the run without an automatic retry.

\end{enumerate}
\vspace{4pt}\hrule
\captionsetup{hypcap=false}
\captionof*{table}{The full construction. The branch-specific update/read path makes a later candidate eligible; it does not by itself certify functional participation or an accepted recursive-improvement chain. Those observations are classified separately using Equation~\ref{eq:functional} and Definition~\ref{def:chain}.}
\end{minipage}
\end{center}

Functional probes bind an actual update/read transition to the candidate-producing history and receive their own resource reservations. The resulting evidence determines whether the admitted versions form the chain defined below.

\Needspace{8\baselineskip}
\begin{definition}[Finite MMLA recursive improvement]
\label{def:chain}
A finite accepted chain $\eta_0\to\eta_1\to\eta_2$ has the proposed recursive dependency when each child is produced through functional participation of its parent adaptation rule, each accepted edge has independent improvement evidence under the declared task-and-cost criterion, and the accepted $\eta_1$ actually participates in generating $\eta_2$. Participation follows the RTT policy path, the RTMA memory path, or their declared joint closure. A claim that both routes participate additionally needs separate path witnesses. Longer chains repeat these obligations within the fixed mutable domain.
\end{definition}
One independently improved and accepted rule demonstrates a useful change to an adaptation method. A second accepted edge, with the first successor now acting in the proposal process, establishes the finite recursive chain. A mere hash difference, a version label in a prompt, or selection by an unchanged external optimizer does not substitute for functional parent participation.

\subsection{Interventions and practical comparisons}
\label{sec:interventions}
For RTT, two interventions address different parts of the dependency. \emph{Disabling the update} replaces $U_\eta$ by the identity on both $\Phi$ and its private state, while holding the reached prefix, arrived feedback, and other inputs fixed. It asks whether numerical updating participates in the next proposal. \emph{Replacing the updater} runs the evolved $U_{\eta_k}$ and original $U_{\eta_0}$ on the same pre-update $(\Phi,s,g,z)$ within their common state space, then compares the subsequent read. It asks whether the evolved rule, beyond merely having an update, changes successor generation. With nonzero $s$ this is a local common-state intervention; a complete trajectory comparison starts both arms from their bound initial states.

For RTMA, disabling the operation forces NULL while preserving the arrived record, state, and budget accounting. Replacing the manager evaluates $V_{\eta_k^M}$ and $V_{\eta_0^M}$ on the same $(M,d)$, then compares the resulting pinned reads and proposal kernels. Changed scores alone are not participation evidence: the selected action and read must matter. A joint probe intervenes on each rule and state path independently, then on the pair. A joint effect alone cannot be assigned to both routes.

These interventions use the action-distribution criterion or the explicitly conditional replay object specified in Section~\ref{sec:rtt}. Fixing one random tape is useful for debugging a transition, but does not silently convert a replay difference into a difference of marginalized kernels. Whole-history probes must also specify which past updates are replaced; changing only the current call and replacing a parent over the entire generation answer different questions.

A separate performance comparison runs complete proposal trajectories under the evolving rule or pair, its original fixed version, and a fixed external search method in the same candidate language and common total budget. It measures whether the evolving process finds useful successors more effectively. Trajectories can diverge after the intervention; the locked outputs receive independent evaluation. Such an advantage concerns discovery quality and economic value, rather than the existence of the recursive structure or the definition of a finite accepted chain.

The external search method receives the same legitimate development information, representation, evaluator, and total allocation, with the same memory/read capacity and archive-access restrictions. Generation, candidate execution, backward passes, failed proposals, audit, and recovery all count. If external search wins, the observed implementation may be less practical while retaining its specified recursive dependency.

\section{Conditional Properties of the Construction}
\label{sec:properties}
The method is the adaptation-rule-to-proposal construction in Section~\ref{sec:recursive}. The following properties describe its causal execution, accepted transitions, and finite resource use under the stated assumptions.

\begin{proposition}[Causal composition]
Suppose each generation begins with an adaptation-rule version measurable with respect to its admitted past; each within-task transition is measurable with respect to arrived records and permitted randomness; and an audit is opened only after the candidate is locked. If callbacks are epoch-bound and cannot modify a closed task, every action in a finite sequence of generations is measurable with respect to the information available at its execution time.
\end{proposition}
\begin{proof}
Condition on the admitted state at the start of a generation. Composition of the causal proposal kernel, arrived-feedback map, and selected adaptation operation yields a causal next state. Induction covers the finite attempts in that generation. Candidate locking precedes the audit; admission is therefore a function of information available at the generation boundary. This state supplies the induction hypothesis for the next generation. Epoch checks exclude a late write from a closed task.\end{proof}

\Needspace{14\baselineskip}
\begin{proposition}[Finite admission on a fixed task distribution]
Let $R(\eta)\in[0,1]$ be the loss of a bound learner using the selected rule or rule pair $\eta$ on a fixed task distribution and resource rule. For every locked candidate and its parent, suppose audit intervals cover their risks simultaneously with probability at least $1-\alpha$. Admit an edge only if
\begin{equation}
 U_k(\eta')+\delta\leq L_k(\eta_k),\qquad\delta>0,
 \label{eq:admit}
\end{equation}
and its deterministic resource checks pass. On the simultaneous coverage event, every accepted edge reduces risk by at least $\delta$; hence there are at most $\lfloor R(\eta_0)/\delta\rfloor$ accepted edges.
\end{proposition}
\begin{proof}
Coverage gives $R(\eta')\leq U_k(\eta')\leq L_k(\eta_k)-\delta\leq R(\eta_k)-\delta$. Sum along accepted edges and use $R\geq0$. A rejected proposal does not change the bound learner's admitted rule, though it consumes resources.\end{proof}
One way to obtain the stated event is to allocate conditional error probabilities to all candidate--parent comparisons, with their sum at most $\alpha$, and use fresh audits with valid conditional intervals. Reusing an ordinary bootstrap interval after repeatedly selecting on the same audit data does not establish this assumption. The component bootstrap intervals reported later are their original finite-study summaries; they are not being promoted to sequential admission guarantees.

\begin{proposition}[Finite work]
If there are at most $K$ generations, at most $N_k$ proposals per generation, a finite maximum instruction count and state size for every adaptation call, and an enforced reservation for every generation, execution, update, and audit operation, total work is bounded by the sum of the reserved costs.
\end{proposition}
\begin{proof}
There are finitely many operations and each has an enforced finite bound. Summing these bounds covers successful and failed operations alike. An operation without a sufficient reservation cannot begin.\end{proof}
A decreasing budget variable alone is insufficient if a loop can perform infinitely many nominally zero-cost operations. The finite generation and instruction bounds are therefore part of the protocol, not only accounting metadata.

The admission result applies to accepted edges on the fixed distribution, and the work bound applies whether candidates are accepted or rejected. Existence of an admissible successor and performance after a distribution change are empirical questions outside those conditions.

\section{Component Evidence and Design Motivation}
\label{sec:evidence}
The completed studies summarized here establish implementations of the numerical carrier and its update, retention, read, and restoration operations. They do not evaluate the new recursive memory-manager domain or a joint recursive chain. They also expose difficulties that a recursive updater must address. Program selection and CI reuse the LARC experiments~\cite{larc}; the SQL studies are subsequent completed evaluations. The four populations and endpoints are distinct. Appendix~\ref{app:component} retains their full results and uncertainty, and Appendix~\ref{app:cost} retains all reported cost scopes. No new experiment was run for this report.

\begin{table}[htbp]
\centering\small
\begin{tabularx}{\linewidth}{@{}lYY@{}}
\toprule Component study & Main observation & Strong comparison or boundary \\
\midrule
Four-program selection & Retention reduces expected error by 24.65 and 36.65 points for static and adapted initializations. & Real-keep error 29.302\% / 24.611\%; support-loss selection 0.78125\%. \\
Chronological CI replay & $G=-0.05337$ while $D=+0.14969$: real binding helps relative to sham, retention harms. & Half-Brier loss rises from 0.12744 to 0.18081; HEDGE4 reaches 0.10243. \\
SParC interaction & Whole-interaction success 18.42\% versus frozen six-proposal 20.18\%; paired gain $-1.75$ points. & 97.5\% interval $[-4.39,0.88]$; strict prefix-read task gains are zero. \\
CoSQL correction transfer & Final-query success 46.965\% versus frozen 47.944\%; paired gain $-0.979$ points. & 97.5\% interval $[-4.247,1.759]$; neither arm passes its online $\Phi$ requirements. \\
\bottomrule
\end{tabularx}
\caption{Existing component evidence, not an evaluation of the proposed recursive routes. Percentage-point differences and half-Brier differences have different units; the studies are not pooled.}
\label{tab:component-summary}
\end{table}

The program-selection result is a development evaluation over 16 parameter groups and three paired training seeds; its final audit remained unopened. It concerns four supplied candidate programs rather than free generation. Both learned initializations benefit from feedback, although a direct same-information rule is much more accurate. Expected and greedy decisions also order the two arms differently, and the fixed stability rule selects the static arm.

CI uses a repository-local retrospective queue, with equal repository weights over one NumPy commit and 71 pandas commits. Its three-seed means are descriptive rather than population intervals. The negative retention effect and positive binding effect show that correctly associated feedback can still drive harmful learning. The later same-batch non-descent diagnostic does not uniquely identify the cause.

The SQL studies provide task-level evaluations with seeds averaged inside tasks and databases as the resampling units. In SParC, saved replay states change some proposal tokens and selected queries without changing the paired task endpoint. CoSQL's trained task models greatly improve over the original substrate (7.295\% final-query success), but their additional online increment fails the registered requirements. The copy-edit, retrieval, and search rules reach 1.824\%, 5.348\%, and 0.961\%; selection remains the development-chosen copy-edit rule. Both SQL endpoints measure reference execution concordance on fixed databases, not universal SQL equivalence. Shared feedback-text and selection paths remain part of their declared reads.

These findings motivate searching over the learning rule while retaining separate measurements of feedback binding, state use, and task benefit. They provide component foundations, rather than a test of the recursive construction's success or failure. A numerical update that helps a task, an updater that improves task learning, and an updater that helps produce a useful successor are observations at different levels.

\section{Relation to Other Self-Improvement Routes}
\label{sec:related}
MMLA-RSI connects policy and memory adaptation with recursive changes to their respective methods. Its distinction from other routes is clearest along three axes: what changes, when feedback becomes a learning signal, and how the changed object participates in the next improvement. These axes overlap; they do not partition all systems into mutually exclusive families.

RRSI regularizes the evolution of an agent harness: prompts, control flow, tools, memory, and context management around a frozen backbone~\cite{rrsi}. The current harness produces traces and feedback, proposal uses the evolution history with a constrained edit budget, and screening and cost-aware selection determine the next incumbent. The inspected algorithms keep the outer proposal and selection rules prescribed. This is an agent-system route to RSI. MMLA-RSI specifies two narrower mutable domains and their proposal-state pathways: the current numerical updater changes proposal $\Phi$, and the current memory manager changes proposal $M$. The RTMA route overlaps memory and context-management search; its defining construction is the explicit reuse of the candidate management rule in managing the evidence that generates its successor.

MemEvolve jointly evolves experiential knowledge and the memory architecture through programmatic encoding, storage, retrieval, and management modules~\cite{memevolve}. Its outer process selects architectures using performance, cost, and latency, diagnoses their execution trajectories, and redesigns permitted modules. It is a direct precedent for improving the memory mechanism itself. The RTMA instance here fixes the representation and reader while varying a bounded retention-and-replacement rule. Equation~\ref{eq:memory-recursion} places that rule in a particular proposal interface: the current manager controls which arrived evaluation records the successor generator can read, and an admitted successor takes over that proposal-memory role in the next generation. The comparison concerns this parent-to-proposal dependency and the mutable domain.

On-policy distillation (OPD) supervises a student's own rollout prefixes with teacher distributions. Agarwal et al.\ develop this approach through Generalized Knowledge Distillation~\cite{gkd}; the Thinking Machines account provides a later practical treatment~\cite{opd}. It describes a sampling and supervision relation; it does not by itself specify updates inside an unresolved answer or changes to the optimizer. In On-Policy Self-Distillation (OPSD), a teacher from the same model family receives privileged context; the reported experiments keep that teacher fixed at the initial policy~\cite{opsd}. The more recent DCE+SRCL framework reuses the updated checkpoint for both student and privileged teacher in the next round~\cite{dce}. This is a real recursive dependence: learned ability changes subsequent supervision. Its specified numerical optimization rule remains fixed. MMLA-RSI's RTT branch changes that numerical rule and carries it into successor generation; its RTMA branch changes evidence retention through a memory-management rule. An OPD objective could be used inside an RTT system if its supervision were legally available at the required time.

The temporal distinction from TTT also needs care. Recent aTTT updates an episode-specific LoRA from trajectory text during a live episode, then uses the adapted state in later turns~\cite{attt}. Its repetition-aware token weighting limits reinforcement of repeated text. It therefore directly overlaps RTT's within-episode numerical update timing. The separate Agentic-TTT work learns a policy for choosing adaptation tools and reusing skills across queries~\cite{agenticttt}. MMLA's RTT account contributes an explicit same-problem state lifetime, read closure, and functional feedback witness; the MMLA-RSI RTT extension additionally uses a changed updater in generating its successor. Whether another implementation satisfies these obligations depends on its actual process, not its acronym.

\begin{table}[htbp]
\centering\small
\begin{tabularx}{\linewidth}{@{}lYY@{}}
\toprule Route & Main changed object & Dependence carried into later improvement \\
\midrule
aTTT & Episode-specific numerical adapter & Updated policy changes later trajectory text under a prescribed update rule. \\
RRSI & Agent harness around a frozen model & Successor harness produces the next round's evidence and evolution history. \\
MemEvolve & Programmable memory architecture & Architecture-dependent execution traces inform selection, diagnosis, and redesign of memory modules. \\
OPSD & Student policy & Student sampling changes; the experimental teacher remains fixed. \\
DCE+SRCL & Student and privileged teacher checkpoints & Updated ability changes the next round's supervision. \\
STOP / DGM & Improver program / self-modifying agent & Successor can act in producing a later descendant. \\
MMLA-RSI (proposed) & Numerical updater, memory manager, or their pair & Successor adapts the proposal policy or memory used to produce the next rule. \\
\bottomrule
\end{tabularx}
\caption{Different realizations of self-improvement. The table identifies demonstrated dependencies in the cited methods and the specified dependency of the proposed MMLA-RSI system. It is not a performance ranking.}
\label{tab:routes}
\end{table}

Self-Refine and Reflexion use feedback to revise outputs or maintain textual reflections~\cite{selfrefine,reflexion}. Such information can improve later attempts without a numerical update. They provide relevant same-information context and memory controls. Retaining a reflection under a fixed manager and recursively changing that manager are distinct operations; both may be useful. Test-time training learns from test inputs~\cite{ttt}, while TTT layers treat a learnable hidden state as a sequence-model component~\cite{tttlayers}. Discover uses reinforcement learning at test time to search for solutions to open problems~\cite{discover}. These works show why learning time, mutable object, and feedback source should be compared separately.

MAML and its first-order approximation learn an initialization from which task adaptation can proceed~\cite{maml,fomaml}. Learned optimizers instead learn an update rule~\cite{learnedopt}. Both are relevant ingredients. Kirsch and Schmidhuber study self-referential meta learning through shared neural functionality, with fitness-monotonic execution allocating computation according to observed performance~\cite{selfreferential}. That work directly addresses learning mechanisms that participate in changing themselves. MMLA-RSI specifies bounded symbolic rules, policy or memory adaptation of the proposal process under those rules, and independent admission before version replacement.

Chen et al.\ formulate optimization-algorithm discovery as program search and discover Lion~\cite{symbolicopt}. Representing and searching update programs is therefore an established approach. In the RTT branch, the current update program additionally trains the proposal policy producing its successor. Its bounded scalar-schedule language gives one concrete domain; the RTMA language gives a separate memory-management domain with the same parent-participation requirement.

STOP is a direct precedent for procedure recursion: its update has the form $I_t=I_{t-1}(\hat u,I_{t-1},L)$, so the current improver produces its successor~\cite{stop}. The authors' qualification that their frozen language model is not full RSI uses a different boundary from improvement-process recursion. Darwin G\"odel Machine retains self-modifying agents in an archive, allowing a descendant to become a parent of a later modification~\cite{dgm}. Its frozen foundation model and fixed archive management do not erase that recursive dependence. Promptbreeder changes mutation prompts and applies them to task-prompt mutation~\cite{promptbreeder}; the inspected method uses an explicit higher-level prompt for mutation-prompt evolution, so its levels should not be collapsed into unrestricted recursion.

G\"odel machines take a different route: self-modification is justified by proofs of utility under a formalized system~\cite{goedel}. Equation~\ref{eq:admit} is an empirical finite-audit condition with stated statistical assumptions, not a replacement for such a proof system or a universal self-improvement theorem. The practical comparison is between the objects that can change, the evidence that admits a change, and the role the changed object plays in the next improvement.

\section{Implementation, Cost, and Reproducibility}
\label{sec:artifacts}
\subsection{A version is a complete computational identity}
A reproducible learner binds the substrate revision, shared offline adaptation, slow factors, updater implementation, tokenizer, numerical configuration, and task reader. Each study uses its own complete computational identity: the program study binds its SFT lineage and always-enabled offline component, while the SQL studies bind their task adapters. Loading verifies the original assets against that identity; detailed version relationships remain in the accompanying lineage records.

Each episode package should save the actual factor tensors, bound slow version, update count, necessary optimizer and random-number state, task and reader identity, and input cursor. Each generation package adds the domain and parent rules, generated candidate programs, selection record, audit split identity, acceptance decision, and cumulative cost. RTMA packages also bind $M_0$, complete committed rows, provenance and predecessor receipts, capacity, row serializer, trusted committer, and pinned read/index identities. Joint packages preserve the exact update interleaving. A functional probe and a successful save/load operation answer different questions and receive separate records.

The RTT, authoritative-memory, and dual-state modules are part of MMLA v4, and the numerical carrier report is available as LARC~\cite{mmla,larc}. The companion source package for this report contains the manuscript, editable vector figures, rendered figure PDFs, bibliography, aggregate figure data, and a manifest binding the completed evidence inputs. The full private research workspace, raw trajectories, and model weights are not included in this writing package. Their availability is separate from the public paper sources. No public URL is asserted for an unpublished recursive implementation.

Detailed component costs, failed-entry charges, and physical/effective work are retained in Appendix~\ref{app:cost}. The same accounting scope applies to the proposed recursive algorithm: discovering, auditing, and executing an adaptation rule or rule pair are charged together, including memory construction, validation, and reading.

\section{Discussion}
\label{sec:discussion}
MMLA-RSI develops two ways for an adaptation method to participate in producing its successor. RTT changes the policy state through numerical learning. RTMA changes the authoritative evidence available to the next attempt. Their recursive layers change the numerical updater and memory manager, respectively, then return the admitted rule to that same role in the next generation. The joint construction retains both state paths instead of treating memory as an accessory to policy learning.

The component evidence identifies practical difficulties: feedback can change behavior without improving it, a strong direct rule can outperform a learned carrier, and offline training can explain most gains attributed to an online system. These observations motivate measurements of complete task performance, state use, and discovery cost. They neither provide nor rule out an effective recursive chain in the proposed domains.

The minimal languages make limited parts of the adaptation process mutable. RTT varies scalar schedules within a fixed numerical template. RTMA varies which complete evidence rows are retained, including the option to write nothing. The trusted memory authority, record semantics, and evaluators stay fixed. More expressive memory transformations or numerical optimizers would be separate domains with their own bounds and comparisons.

The empirical question is whether a rule admitted for improved task behavior then helps produce another admitted rule, and whether that discovery process is useful relative to fixed rules and external search at the same information and total cost. Either route can be evaluated first. A joint claim additionally measures the complete task and both state paths, including the cost of invalidating or recomputing memory values when the reader changes.

\appendix
\section{Illustrative RTT and RTMA Generation Traces}
\label{app:example}
Consider a legal member of the proposed updater language with fixed tensor shapes:
\begin{align}
 \tilde g_j&=g_j\min\{1,\kappa/(\|g_j\|+\varepsilon)\},\\
 s_{j+1}&=\operatorname{clip}_{[-S_{\max},S_{\max}]}\!\left(\mu s_j+(1-\mu)\tilde g_j\right),\qquad
 \Delta\Phi_j=-a_j s_{j+1}.
\end{align}
Here the three expression trees return bounded $a_j,\kappa,\mu$ as specified in Section~\ref{sec:recursive}, and the factor update uses its declared projection. This instance changes the scalar schedules within the fixed clipped-momentum template. A comparison fixes the initial factors and optimizer state so that its effect is attributable to the updater.

In generation zero, the proposal learner starts from $\rho^{\rm prop}$ and samples a first program. Its development execution returns a reward; Equation~\ref{eq:proposalgrad} gives a gradient; $U_{\eta_0}$ updates the proposal state. The next sampled program is therefore a fresh attempt generated after the update. After a bounded number of attempts, development selection locks one candidate. Independent audit may admit it as $\eta_1$. Generation one resets proposal factors to the same bound initialization and now uses $U_{\eta_1}$ to process proposal feedback. If a resulting $\eta_2$ is independently better and parent participation is verified, the trace contains the finite recursive chain defined in Section~\ref{sec:recursive}.

This trace requires two distinct uses of candidate evaluation. Development executions teach and select; audit executions estimate utility after selection. A successful audit answer cannot be silently fed back into the proposal trajectory whose independence it was meant to establish. It can enter the declared history of a later generation, which then needs fresh audit tasks.

\subsection*{A memory-management trace}
Take a capacity-two proposal memory, initially empty. When the fixed syntax bounds admit it, one legal score tree is
\[
 e_M(f)=(1-f_1)\bigl[(1-f_2)(1-f_3)+f_2(f_5-f_3)\bigr].
\]
It uses only the existing features, constants $0,1$, and arithmetic operations. NULL scores zero; insertion into an empty slot scores $1-\ell$; replacement of an occupied slot scores $\ell_i-\ell$. With NULL winning ties, this illustrative rule retains lower-loss records when their score is positive and otherwise leaves memory unchanged. It is a concrete program in the stated language, not a measured result or a claim that low-loss records always have greater future value. A first candidate with development feedback loss $\ell<1$ supplies a complete provenance-bound row. The parent manager selects an empty slot; the trusted committer installs the row, and the next proposal reads its exact contents. After later evaluations fill the slots, a write can replace an older record or select NULL. The generator receives only the header and current rows, so evicted evidence does not persist in a separate prompt history.

A candidate successor manager is executed on fresh development task memories. If selected and independently admitted, it becomes the manager of the next generation's proposal memory, starting again from the same $M_0^{\rm prop}$. Its different scoring program can retain a different set of evaluation records and thereby affect the successor proposals. This describes the specified dependency, not an observed beneficial run. With joint rules, the same arrived evaluations also update the proposal residual through the parent numerical updater, with both pinned versions recorded.

A parent manager that always selects NULL, such as the all-zero score tree, makes no non-NULL commit. RTMA-only and joint runs then have no eligible successor under step~6 of Algorithm~\ref{alg:recursive} and terminate, even if an independently sampled program has better task performance. Thus a legal initial rule need not provide the state transition required to start a recursive chain.

\section{Detailed Component Evidence}
\label{app:component}
We use completed aggregate artifacts to examine what the existing learning state can do and where the proposed recursive layer would need to help. The program-selection and continuous-integration (CI) studies are reused from the LARC report~\cite{larc}; the two SQL studies are subsequent completed task-level evaluations. They have different populations, read closures, and endpoints. Their scores are not pooled. No result from the separate evaluation still running at the evidence cutoff is included.

\subsection{Program selection: useful adaptation with a stronger direct rule}
The program task presents four complete candidate programs and supplies their execution feedback on support examples. The learner selects a program; it does not freely generate one. Both training objectives use the same input LARC4 position on the bound MiniCPM5-1B-SFT substrate with the shared offline component enabled. Two support-gradient steps use SGD with step size $0.1$, no momentum, no weight decay, and no inner clipping. The numerical query predicts a distribution over the four candidates, and expected execution error is the primary endpoint.

The saved development evaluation covers 16 parameter groups, four recipes per group, and three paired training seeds. Its 384 repeated rows are not 384 independent tasks. The reported 99\% intervals resample parameter groups within the declared family strata, conditional on those three trained seeds. The separate final audit remained unopened in this study.

\begin{table}[htbp]
\centering\small
\begin{tabular}{@{}lrrrr@{}}
\toprule Initialization & Real keep & Real reset & Sham keep & Sham reset \\
\midrule
Static objective & 29.302 & 53.948 & 57.666 & 53.948 \\
Adapted objective & 24.611 & 61.259 & 60.379 & 61.259 \\
\bottomrule
\end{tabular}
\caption{Expected program-selection error (\%, lower is better). Reset restores each arm's own trained initialization. The support-loss selector reaches 0.78125\% error.}
\label{tab:programcells}
\end{table}
\begin{table}[htbp]
\centering\small
\begin{tabular}{@{}lrr@{}}
\toprule Contrast & Mean (percentage points) & 99\% group interval \\
\midrule
Static retention $G$ & 24.65 & [19.39, 29.26] \\
Static binding $D$ & 28.36 & [21.85, 34.93] \\
Adapted retention $G$ & 36.65 & [31.65, 41.20] \\
Adapted binding $D$ & 35.77 & [30.33, 41.01] \\
Static error $-$ adapted error, real keep & 4.69 & [1.19, 8.88] \\
\bottomrule
\end{tabular}
\caption{Within-arm feedback effects and between-objective increment answer separate questions. Intervals are conditional on the three paired training seeds.}
\label{tab:programcontrasts}
\end{table}

Both arms benefit from real-feedback retention. The adapted objective has the lower mean post-update error, while its between-objective differences vary across seeds: $+1.45$, $-10.46$, and $+23.09$ percentage points in its favor. The predeclared stability rule selects the static initialization because its worst-seed real-feedback gain is 20.16 points, versus 18.64 for the adapted initialization. Selection therefore preserves the distinction between mean objective superiority and stability across the three trained instances.

The larger adaptation increment of the adapted arm also has two sources. Its $36.65-24.65=12.00$ point larger $G$ decomposes into a 7.31 point worse reset error and a 4.69 point better retained error. Reporting only the 12.00 point contrast would conflate a worse starting point with a better final policy. Greedy decisions tell another part of the story: static keep has 15.49\% greedy error and adapted keep has 16.04\%, despite the ordering of their expected errors. Appendix~\ref{app:secondary} preserves the corresponding secondary table.

A direct rule evaluates each candidate's support loss and selects the best. Its 0.78125\% error is far below either numerical learner's error. Thus this study establishes useful feedback adaptation within the specified learner, alongside a much stronger same-information rule. It does not establish an advantage in open-ended program synthesis or a reason to deploy the learner instead of the rule.

\subsection{CI: feedback binding and retention can disagree}
The CI study replays public job outcomes in repository-local chronological order. It contains 3,321 jobs across 72 commits: 83 NumPy jobs from one commit and 3,238 pandas jobs from 71 commits. Jobs are aggregated to commits, commits are weighted equally within each repository, and the two repositories are weighted equally. Consequently the single NumPy commit has 71 times the weight of one pandas commit. This is a retrospective repository-balanced replay, not a global prospective deployment sample.

The score is half-Brier loss~\cite{brier},
\begin{equation}
 \ell_{\rm HB}(p,y)=\frac12\sum_c(p_c-\mathbf1\{y=c\})^2.
 \label{eq:brier}
\end{equation}
HEDGE4 is this study's four-expert exponential-weights combination, based on the HEDGE framework~\cite{hedge}. The name does not import a delayed-feedback regret guarantee for this implementation.

\begin{table}[htbp]
\centering\small
\begin{tabular}{@{}lr@{}}
\toprule Method, with recorded memory condition & Mean half-Brier loss \\
\midrule
Shared real/sham reset read, real $M$ & 0.12744 \\
Real feedback, keep $\Phi$, real $M$ & 0.18081 \\
Permuted feedback, keep $\Phi$, real $M$ & 0.33050 \\
HEDGE4 & 0.10243 \\
\midrule
Real retention $G$ & $-0.05337$ \\
Feedback binding $D$ & $+0.14969$ \\
\bottomrule
\end{tabular}
\caption{CI replay. The three-seed means are descriptive summaries on two recorded repositories, not population confidence intervals. HEDGE4 has the lowest mean loss among the methods shown.}
\label{tab:ci}
\end{table}
Both logical reset cells use the same P0M1 prediction, evaluated from each seed's task-trained initialization $\rho$ and the same real-memory prompt. The permutation changes only fast-update labels; its private random-number state is not read by the probability scorer. The implementation evaluates this common reset read once, rather than separately rolling back and reevaluating both updated branches. Thus the registered contrast is
\[
 D=(L_{\rm reset}-L_{\rm real,keep})-(L_{\rm reset}-L_{\rm sham,keep})
   =L_{\rm sham,keep}-L_{\rm real,keep}\approx0.14969.
\]
The equality concerns the per-job prediction readout under fixed real memory. The update histories, private random-number states, and subsequent learning trajectories need not coincide.

Retaining real updates is worse than resetting in every seed, while real binding is better than permuted binding in every seed. With $M$ disabled, the mean retention effect remains negative, $-0.04400$. Correct feedback binding therefore does not establish useful online learning. A subsequent fixed diagnostic in the LARC study records same-batch cross-entropy non-descent and inconsistent future benefit from shrinking updates; it does not uniquely identify a single root cause~\cite{larc}.

This experiment adapts across jobs. It informs the stability of a numerical learning carrier and the interpretation of $G$ and $D$, but it is not a same-unresolved-problem RTT witness. The unit and timing of feedback matter as much as the existence of a backward pass.

\begin{figure}[htbp]
\centering\includegraphics[width=\linewidth]{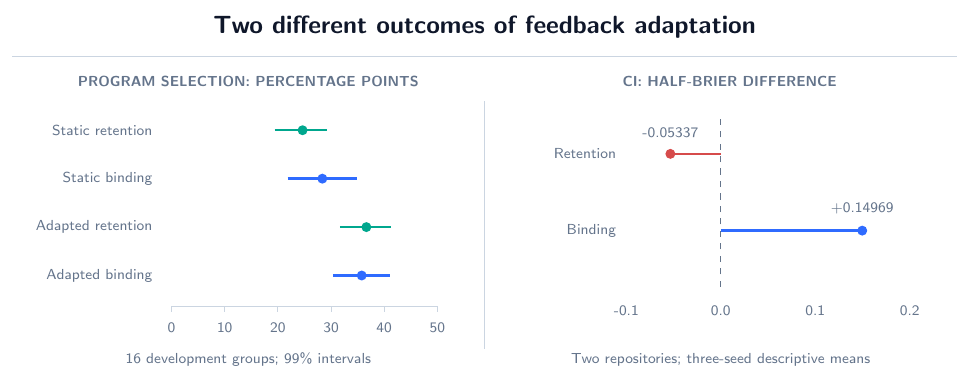}
\caption{\textbf{Binding is not the same as benefit.} Left: program-selection retention and binding effects with original 99\% development-group intervals. Right: CI effects in half-Brier units, shown as three-seed means without population intervals. Positive values favor the named effect in both panels; the units and populations differ.}
\label{fig:evidence}
\end{figure}

\subsection{SQL interaction: state changes without endpoint gains}
The first completed task-level SQL study uses grouped interactions from SParC~\cite{sparc}, with 56 development interactions across 28 databases and 75 confirmation interactions across 19 databases. Its primary endpoint is whole-interaction success: every turn must produce a query execution agreeing with the reference on the fixed database. This is an operational reference-concordance score, rather than a proof that two SQL programs are equivalent on all databases. Seeds are averaged within the interaction and then databases are weighted equally.

The adapted real-feedback method reaches 18.42\% confirmation success, compared with 20.18\% for the static six-proposal frozen method. The paired difference is $-1.75$ percentage points, with a 97.5\% database-bootstrap interval of $[-4.39,0.88]$. The three registered non-neural rule methods each score zero on this confirmation population; the development-selected rule remains the comparator. Beating those particular rules does not establish an advantage over general-purpose SQL systems.

The saved-state audit is informative even though the main comparison fails. Across development and confirmation, 393 paired cases contain 1,965 fast-state packages. The audit verifies nonnull gradients and state changes in 786 keep packages and exact initialization, optimizer, and random-state restoration in 786 reset packages. On 225 confirmation task--seed pairs, real keep versus reset changes 14 third-proposal token sequences and three selected SQL outputs, while whole-interaction and last-turn scores remain unchanged in every paired case. Real versus sham produces different actual rewards and fast states in only six of these 225 pairs. This is evidence of local saved-output sensitivity and weak effective sham separation, not positive task benefit.

The offline SQL adaptation includes a rank-16 task adapter and the rank-4 input residual, totaling 700,416 trainable entries. Gains over the original model cannot be attributed to the fast residual alone. The strict saved-prefix retention and binding contrasts are zero at the task endpoints despite the verified state and output changes.

\subsection{CoSQL correction transfer: online increment below the prespecified requirements}
The next completed study uses CoSQL~\cite{cosql} under its own fixed correction-transfer protocol. The evaluation has 358 development episodes across 28 databases and 210 confirmation episodes across 19 databases, with three paired seeds and 43 method--seed cells over 568 episodes. All 24,424 logical records are complete. The endpoint is final-query execution concordance with the reference, macro-averaged by database. It is distinct from the previous study's whole-interaction success and is not presented as an official benchmark leaderboard score.

\begin{table}[htbp]
\centering\small
\begin{tabular}{@{}lr@{}}
\toprule Confirmation method & Final-query success (\%) \\
\midrule
Original frozen substrate, six proposals & 7.295 \\
Development-selected copy-edit rule & 1.824 \\
Retrieval rule & 5.348 \\
Search rule & 0.961 \\
Static task model, six frozen proposals & 47.944 \\
Static task model, real-feedback procedure & 48.230 \\
Adapted task model, real-feedback procedure & 46.965 \\
\bottomrule
\end{tabular}
\caption{Completed CoSQL correction-transfer evaluation. The copy-edit rule remains the selected rule even though retrieval scores higher on confirmation. The current, separate contrast-training evaluation is not included.}
\label{tab:cosql}
\end{table}
The adapted real-feedback procedure differs from the static frozen comparator by $-0.979$ percentage points, with a 97.5\% interval of $[-4.247,1.759]$. The registered task-level gate requires at least five points and a positive 97.5\% lower bound against both the frozen method and the development-selected rule. It fails against the frozen method.

\begin{table}[htbp]
\centering\small
\begin{tabular}{@{}lrr@{}}
\toprule Arm and controlled effect & Mean (percentage points) & 95\% database interval \\
\midrule
Static retention & 0.775 & $[-0.638,2.930]$ \\
Static binding & 0.387 & $[-0.196,0.971]$ \\
Adapted retention & 0.580 & $[0.000,1.565]$ \\
Adapted binding & 0.423 & $[-0.125,1.260]$ \\
\bottomrule
\end{tabular}
\caption{CoSQL controlled fast-state effects. Each arm requires both binding and retention lower bounds strictly above zero; neither meets that requirement. The strict checks use the stored unrounded bounds; the adapted-retention lower bound is exactly zero in the saved aggregate. These paired prefix reads differ from the complete procedures in Table~\ref{tab:cosql}.}
\label{tab:cosqlphi}
\end{table}
The final audit checks 13,632 prefix bindings and 3,408 reset/query-closure cases. Memory is disabled, but shared support text, proposal history, feedback, tools, and selectors remain in the declared read. The results therefore concern the additional numerical effect conditional on those channels. The substantial improvement over the original frozen substrate demonstrates a better task model. The prespecified deployment requirements---the task endpoint and the online $\Phi$ requirements---remain unmet for this numerical recipe; the completed models and failed comparisons are retained.

\section{Measured Costs and Accounting Scope}
\label{app:cost}
\begin{table}[htbp]
\centering\small
\begin{tabularx}{\linewidth}{@{}lYrr@{}}
\toprule Study & Accounting scope & Recorded quantity & Value \\
\midrule
Program selection & Full entry, including production preflight & Wall seconds & 1,634.50 \\
 & Contract charge, including data work & Seconds & 1,640.27 \\
 & Physical / effective token positions & Millions & 49.026 / 19.521 \\
CI & Qualification, task fit, and repeated evaluation & Native wall hours & 3.675 \\
SQL interaction R5 & This successful R5 invocation & GPU residency upper hours & 4.819 \\
 & Budgeted process-tree CPU & CPU hours & 5.641 \\
 & Native wall snapshot & Wall hours & 5.266 \\
CoSQL transfer & Cumulative finite recipe, including failed work & GPU residency hours & 14.359 \\
 & Cumulative process-tree CPU & CPU hours & 21.739 \\
 & Original origin, including pauses and review & Wall hours & 24.888 \\
\bottomrule
\end{tabularx}
\caption{Recorded component costs. Wall time, process-tree CPU, GPU residency, and effective computation are separate quantities. They must not be summed as though they were one resource measure.}
\label{tab:cost}
\end{table}

Program-selection physical positions decompose as $49{,}025{,}792=47{,}435{,}776$ mixer positions $+1{,}585{,}920$ selected-head positions $+4{,}096$ native positions. Its 19,521,408 effective positions count a different computational quantity. The entry includes 1,536 formal outer updates and two production-preflight outer updates; it is not simply the nominal training matrix multiplied by a forward cost.

The CI total includes a 57.84-second failed qualification, 6,626.08 seconds of task-fit work, and 6,545.88 seconds of repeated evaluation. The SQL R5 cost does not include the preceding four failed entries, whose separate GPU residency totals 2,760.68 seconds and CPU budget charges total 3,541.26 seconds. The CoSQL cumulative record includes failed work and the redone third-seed evaluation; its separate result-collection call uses 24.02 CPU seconds and 23.97 wall seconds. These boundaries preserve the actual cost of arriving at a result without folding unrelated experiments into it.

The recursive design must use the same discipline. A fast updater can be expensive to discover; a model that improves after more attempts may lose under an equal complete-task budget. Proposal development, audit evaluation, optimizer computation, and restoration costs all belong to the recursive process.

\FloatBarrier
\Needspace{24\baselineskip}
\section{Secondary Program Decisions and CI Weighting}
\label{app:secondary}
\noindent\begin{minipage}{\linewidth}
\captionsetup{type=table,hypcap=false}
\centering\small
\begin{tabular}{@{}lrrr@{}}
\toprule Program arm/read & Expected error (\%) & Greedy error (\%) & All-query pass (\%) \\
\midrule
Static reset & 53.95 & 47.76 & 42.71 \\
Static keep & 29.30 & 15.49 & 81.77 \\
Adapted reset & 61.26 & 61.72 & 30.21 \\
Adapted keep & 24.61 & 16.04 & 80.73 \\
\bottomrule
\end{tabular}
\caption{Saved secondary program-selection summaries, reused from LARC. A probability-weighted endpoint and a greedy decision need not order the trained arms in the same way.}
\end{minipage}\par\medskip
\noindent\begin{minipage}{\linewidth}
\captionsetup{type=table,hypcap=false}
\centering\small
\begin{tabular}{@{}lrrr@{}}
\toprule Repository & Jobs & Commits & Weight per commit \\
\midrule
NumPy & 83 & 1 & $1/2$ \\
pandas & 3,238 & 71 & $1/142$ \\
\bottomrule
\end{tabular}
\caption{CI population and aggregation. The 71-fold difference in commit weight follows directly from equal repository weighting. Jobs inside each commit contribute to that commit's score.}
\end{minipage}\par\medskip

\section{Evidence Identities and Cutoff}
\label{app:identities}
The manuscript draws its theory from the local MMLA v4 source, including its RTT, authoritative-memory, and dual-state modules; the public citation is fixed at arXiv:2606.28876v4. The local report source has SHA256
\begin{center}\footnotesize\url{d4f24305dcd000a203dcbe24a4dcdb12494a7a4de5dea3eb06ebb1a1c91ca899}\end{center}
The locally supplied LARC report source has SHA256
\begin{center}\footnotesize\url{f7ccfe0af95168f2e4f40e208b3033f842f12c0661b844e661e39711be03ab9a}\end{center}
and the public citation is arXiv:2609.40063v1. A local source identity is not a claim of byte identity with an arXiv-generated PDF. These report sources and the versioned aggregate records fix what was available for this synthesis.

The program-selection substrate is \texttt{openbmb/MiniCPM5-1B-SFT}, fixed at revision
\begin{center}\footnotesize\url{a60b37f1fc409c54e1e337b0723aaac6f92dfec0}\end{center}
Its $\theta_0$ tensor digest and always-enabled $\omega$ byte and tensor digests are preserved in the accompanying \texttt{model-lineage.json}. That identity refers to the program-selection realization; it does not replace the separate task adapters of the SQL studies. The original assets must be checked on load rather than reconstructed by substitute training.

The completed SQL interaction result audit binds candidate
\begin{center}\footnotesize\url{c4c1f33c66d329ad41cb52ac4edbf58d2a686ec2}\end{center}
and the completed CoSQL correction-transfer result audit binds candidate
\begin{center}\footnotesize\url{16c348c5865ac9ee3c200ed43a026788d09dac09daef265a1cd89b91afafbed3}\end{center}
The latter covers the closed correction-transfer recipe, not the separate contrast-training candidate still under evaluation at this report's cutoff. A source review, numerical save/restore check, statistical result review, and observed scientific effect remain separate forms of evidence.

The companion aggregate data preserve the values used to render the figures, and the evidence manifest records original relative paths, byte sizes, and SHA256 hashes. The source archive is self-contained for document compilation. Access to private experiment inputs and trajectories remains subject to their existing access scope; no new public release of those materials is implied by this report.

\FloatBarrier
\begingroup
\small
\bibliographystyle{plain}
\bibliography{references}

\begin{thebibliography}{10}

\bibitem{gkd}
Rishabh Agarwal, Nino Vieillard, Yongchao Zhou, Piotr Stanczyk, Sabela Ramos,
  Matthieu Geist, and Olivier Bachem.
\newblock {On-Policy Distillation of Language Models: Learning from
  Self-Generated Mistakes}.
\newblock \href{https://arxiv.org/abs/2306.13649v3}{arXiv:2306.13649v3}, 2024.
\newblock Version fixed at 2024-01-17.

\bibitem{learnedopt}
Marcin Andrychowicz, Misha Denil, Sergio Gomez, Matthew~W. Hoffman, David Pfau,
  Tom Schaul, Brendan Shillingford, and Nando de~Freitas.
\newblock {Learning to learn by gradient descent by gradient descent}.
\newblock \href{https://arxiv.org/abs/1606.04474v2}{arXiv:1606.04474v2}, 2016.
\newblock Version fixed at 2016-11-30.

\bibitem{brier}
Glenn~W. Brier.
\newblock {Verification of Forecasts Expressed in Terms of Probability}.
\newblock {\em Monthly Weather Review}, 78(1):1--3, 1950.
\newblock
  \url{https://doi.org/10.1175/1520-0493(1950)078<0001:VOFEIT>2.0.CO;2}.

\bibitem{symbolicopt}
Xiangning Chen, Chen Liang, Da~Huang, Esteban Real, Kaiyuan Wang, Yao Liu, Hieu
  Pham, Xuanyi Dong, Thang Luong, Cho-Jui Hsieh, Yifeng Lu, and Quoc~V. Le.
\newblock {Symbolic Discovery of Optimization Algorithms}.
\newblock \href{https://arxiv.org/abs/2302.06675v4}{arXiv:2302.06675v4}, 2023.
\newblock Version fixed at 2023-05-08.

\bibitem{promptbreeder}
Chrisantha Fernando, Dylan Banarse, Henryk Michalewski, Simon Osindero, and Tim
  Rocktäschel.
\newblock {Promptbreeder: Self-Referential Self-Improvement Via Prompt
  Evolution}.
\newblock \href{https://arxiv.org/abs/2309.16797v1}{arXiv:2309.16797v1}, 2023.
\newblock Version fixed at 2023-09-28.

\bibitem{maml}
Chelsea Finn, Pieter Abbeel, and Sergey Levine.
\newblock {Model-Agnostic Meta-Learning for Fast Adaptation of Deep Networks}.
\newblock \href{https://arxiv.org/abs/1703.03400v3}{arXiv:1703.03400v3}, 2017.
\newblock Version fixed at 2017-07-18.

\bibitem{hedge}
Yoav Freund and Robert~E. Schapire.
\newblock {A Decision-Theoretic Generalization of On-Line Learning and an
  Application to Boosting}.
\newblock {\em Journal of Computer and System Sciences}, 55(1):119--139, 1997.
\newblock \url{https://doi.org/10.1006/jcss.1997.1504}.

\bibitem{lora}
Edward~J. Hu, Yelong Shen, Phillip Wallis, Zeyuan Allen-Zhu, Yuanzhi Li, Shean
  Wang, Lu~Wang, and Weizhu Chen.
\newblock {LoRA: Low-Rank Adaptation of Large Language Models}.
\newblock \href{https://arxiv.org/abs/2106.09685v2}{arXiv:2106.09685v2}, 2021.
\newblock Version fixed at 2021-10-16.

\bibitem{selfreferential}
Louis Kirsch and J{\"u}rgen Schmidhuber.
\newblock {Eliminating Meta Optimization Through Self-Referential Meta
  Learning}.
\newblock \href{https://arxiv.org/abs/2212.14392v1}{arXiv:2212.14392v1}, 2022.
\newblock Version fixed at 2022-12-29.

\bibitem{agenticttt}
Jiahao Lu and Mohan Kankanhalli.
\newblock {Agentic-TTT: Training test-time policy for test-time training}.
\newblock \href{https://arxiv.org/abs/2610.12002v1}{arXiv:2610.12002v1}, 2026.
\newblock Version fixed at 2026-10-08.

\bibitem{opd}
Kevin Lu and {Thinking Machines Lab}.
\newblock {On-Policy Distillation}.
\newblock {\em Thinking Machines Lab: Connectionism}, 2025.
\newblock October 2025;
  \href{https://doi.org/10.64434/tml.20251026}{doi:10.64434/tml.20251026};
  accessed October 14, 2026.

\bibitem{selfrefine}
Aman Madaan, Niket Tandon, Prakhar Gupta, Skyler Hallinan, Luyu Gao, Sarah
  Wiegreffe, Uri Alon, Nouha Dziri, Shrimai Prabhumoye, Yiming Yang, Shashank
  Gupta, Bodhisattwa~Prasad Majumder, Katherine Hermann, Sean Welleck, Amir
  Yazdanbakhsh, and Peter Clark.
\newblock {Self-Refine: Iterative Refinement with Self-Feedback}.
\newblock \href{https://arxiv.org/abs/2303.17651v2}{arXiv:2303.17651v2}, 2023.
\newblock Version fixed at 2023-05-25.

\bibitem{fomaml}
Alex Nichol, Joshua Achiam, and John Schulman.
\newblock {On First-Order Meta-Learning Algorithms}.
\newblock \href{https://arxiv.org/abs/1803.02999v3}{arXiv:1803.02999v3}, 2018.
\newblock Version fixed at 2018-10-22.

\bibitem{goedel}
Juergen Schmidhuber.
\newblock {Goedel Machines: Self-Referential Universal Problem Solvers Making
  Provably Optimal Self-Improvements}.
\newblock \href{https://arxiv.org/abs/cs/0309048v5}{arXiv:cs/0309048v5}, 2006.
\newblock Version fixed at 2006-12-17.

\bibitem{reflexion}
Noah Shinn, Federico Cassano, Edward Berman, Ashwin Gopinath, Karthik
  Narasimhan, and Shunyu Yao.
\newblock {Reflexion: Language Agents with Verbal Reinforcement Learning}.
\newblock \href{https://arxiv.org/abs/2303.11366v4}{arXiv:2303.11366v4}, 2023.
\newblock Version fixed at 2023-10-10.

\bibitem{tttlayers}
Yu~Sun, Xinhao Li, Karan Dalal, Jiarui Xu, Arjun Vikram, Genghan Zhang, Yann
  Dubois, Xinlei Chen, Xiaolong Wang, Sanmi Koyejo, Tatsunori Hashimoto, and
  Carlos Guestrin.
\newblock {Learning to (Learn at Test Time): RNNs with Expressive Hidden
  States}.
\newblock \href{https://arxiv.org/abs/2407.04620v4}{arXiv:2407.04620v4}, 2025.
\newblock Version fixed at 2025-08-31.

\bibitem{ttt}
Yu~Sun, Xiaolong Wang, Zhuang Liu, John Miller, Alexei~A. Efros, and Moritz
  Hardt.
\newblock {Test-Time Training with Self-Supervision for Generalization under
  Distribution Shifts}.
\newblock \href{https://arxiv.org/abs/1909.13231v3}{arXiv:1909.13231v3}, 2020.
\newblock Version fixed at 2020-07-01.

\bibitem{attt}
Yanbo Wang, Jinhua Hao, Yuze Shi, Kun Yuan, and Ming Sun.
\newblock {No Time Like the Present: Agentic Test-Time Training for LLM
  Agents}.
\newblock \href{https://arxiv.org/abs/2607.03441v1}{arXiv:2607.03441v1}, 2026.
\newblock Version fixed at 2026-07-03.

\bibitem{williams}
Ronald~J. Williams.
\newblock {Simple statistical gradient-following algorithms for connectionist
  reinforcement learning}.
\newblock {\em Machine Learning}, 8(3--4):229--256, 1992.

\bibitem{rrsi}
Peng Xia, Rujun Han, Zifeng Wang, Yanfei Chen, Yufan Zhuang, Yoonho Lee,
  Chengsong Huang, Han Yu, Zhongying CuiZhu, Yifei Ming, Huaxiu Yao, Burak
  Gokturk, Tomas Pfister, and Chen-Yu Lee.
\newblock {RRSI: Regularized Recursive Self-Improvement of Agent Harnesses}.
\newblock \href{https://arxiv.org/abs/2609.24972v3}{arXiv:2609.24972v3}, 2026.
\newblock Version fixed at 2026-10-04.

\bibitem{dce}
Shangjian Yin, Zehao Zhao, Kavosh Asadi, Rui Liu, Yuchen Lu, Shike Mei, Hang
  Cui, Luke Simon, Zhouxing Shi, and Hamed Firooz.
\newblock {Recursive Self-Improvement via On-Policy Distillation for
  Reasoning}.
\newblock \href{https://arxiv.org/abs/2609.30652v1}{arXiv:2609.30652v1}, 2026.
\newblock Version fixed at 2026-09-25.

\bibitem{cosql}
Tao Yu, Rui Zhang, He~Yang Er, Suyi Li, Eric Xue, Bo~Pang, Xi~Victoria Lin,
  Yi~Chern Tan, Tianze Shi, Zihan Li, Youxuan Jiang, Michihiro Yasunaga,
  Sungrok Shim, Tao Chen, Alexander Fabbri, Zifan Li, Luyao Chen, Yuwen Zhang,
  Shreya Dixit, Vincent Zhang, Caiming Xiong, Richard Socher, Walter~S Lasecki,
  and Dragomir Radev.
\newblock {CoSQL: A Conversational Text-to-SQL Challenge Towards Cross-Domain
  Natural Language Interfaces to Databases}.
\newblock \href{https://arxiv.org/abs/1909.05378v1}{arXiv:1909.05378v1}, 2019.
\newblock Version fixed at 2019-09-11.

\bibitem{sparc}
Tao Yu, Rui Zhang, Michihiro Yasunaga, Yi~Chern Tan, Xi~Victoria Lin, Suyi Li,
  Heyang Er, Irene Li, Bo~Pang, Tao Chen, Emily Ji, Shreya Dixit, David
  Proctor, Sungrok Shim, Jonathan Kraft, Vincent Zhang, Caiming Xiong, Richard
  Socher, and Dragomir Radev.
\newblock {SParC: Cross-Domain Semantic Parsing in Context}.
\newblock \href{https://arxiv.org/abs/1906.02285v1}{arXiv:1906.02285v1}, 2019.
\newblock Version fixed at 2019-06-05.

\bibitem{discover}
Mert Yuksekgonul, Daniel Koceja, Xinhao Li, Federico Bianchi, Jed McCaleb,
  Xiaolong Wang, Jan Kautz, Yejin Choi, James Zou, Carlos Guestrin, and Yu~Sun.
\newblock {Learning to Discover at Test Time}.
\newblock \href{https://arxiv.org/abs/2601.16175v2}{arXiv:2601.16175v2}, 2026.
\newblock Version fixed at 2026-02-05.

\bibitem{stop}
Eric Zelikman, Eliana Lorch, Lester Mackey, and Adam~Tauman Kalai.
\newblock {Self-Taught Optimizer (STOP): Recursively Self-Improving Code
  Generation}.
\newblock \href{https://arxiv.org/abs/2310.02304v3}{arXiv:2310.02304v3}, 2024.
\newblock Version fixed at 2024-08-16.

\bibitem{memevolve}
Guibin Zhang, Haotian Ren, Chong Zhan, Zhenhong Zhou, Junhao Wang, He~Zhu,
  Wangchunshu Zhou, and Shuicheng Yan.
\newblock {MemEvolve: Meta-Evolution of Agent Memory Systems}.
\newblock \href{https://arxiv.org/abs/2512.18746v1}{arXiv:2512.18746v1}, 2025.
\newblock Version fixed at 2025-12-21.

\bibitem{dgm}
Jenny Zhang, Shengran Hu, Cong Lu, Robert Lange, and Jeff Clune.
\newblock {Darwin G\"odel Machine: Open-Ended Evolution of Self-Improving
  Agents}.
\newblock \href{https://arxiv.org/abs/2505.22954v3}{arXiv:2505.22954v3}, 2026.
\newblock Version fixed at 2026-03-12.

\bibitem{opsd}
Siyan Zhao, Zhihui Xie, Mengchen Liu, Jing Huang, Guan Pang, Feiyu Chen, and
  Aditya Grover.
\newblock {Self-Distilled Reasoner: On-Policy Self-Distillation for Large
  Language Models}.
\newblock \href{https://arxiv.org/abs/2601.18734v3}{arXiv:2601.18734v3}, 2026.
\newblock Version fixed at 2026-03-20.

\bibitem{larc}
Junyi Zou and Avrova Donz.
\newblock {LARC: Low-Rank Adaptive Residual Connections for Learning in Frozen
  Models}.
\newblock \href{https://arxiv.org/abs/2609.40063v1}{arXiv:2609.40063v1}, 2026.
\newblock Version fixed at 2026-09-30.

\bibitem{mmla}
Junyi Zou and Avrova Donz.
\newblock {MMLA: Memory-Mediated Learning Architecture for Predictive
  Dual-State Adaptation}.
\newblock \href{https://arxiv.org/abs/2606.28876v4}{arXiv:2606.28876v4}, 2026.
\newblock Version fixed at 2026-09-14.

\end{thebibliography}
\endgroup

\end{document}